\documentclass[journal]{IEEEtran}

\ifCLASSINFOpdf

\usepackage{graphicx}
\usepackage{tikz}
\usepackage{epstopdf}
\usepackage{array}
\usepackage{color,soul}
\usepackage{amsmath}
\usepackage{xcolor}
\usepackage{adjustbox}
\usepackage{multirow}

\else
\fi
\begin{document}
	%
	\title{Unsupervised Feature Learning Toward a Real-time Vehicle Make and Model Recognition}
	
	\author{Amir~Nazemi,
		Mohammad~Javad~Shafiee,~Zohreh~Azimifar~and~Alexander Wong
		\thanks{Amir Nazemi is with the School of Electrical \& Computer Engineering, Shiraz University, Shiraz, Iran (e-mail: nazemi@cse.shirazu.ac.ir).}
		\thanks{Mohammad Javad Shafiee is with the Systems Design Engineering Dept., University of Waterloo, Waterloo, Canada (e-mail: mjshafiee@uwaterloo.ca).}
		\thanks{Zohreh Azimifar is with the School of Electrical \& Computer Engineering, Shiraz University, Shiraz, Iran (e-mail: azimifar@cse.shirazu.ac.ir). }
		\thanks{Alexander Wong is with the Systems Design Engineering Dept., University of Waterloo, Waterloo, Canada (e-mail: a28wong@engmail.uwaterloo.ca).}
	}
	
	%
	%

	\markboth{}
	{Shell \MakeLowercase{\textit{et al.}}: Bare Demo of IEEEtran.cls for Journals}

	\maketitle
	
	\begin{abstract}
		Vehicle Make and Model Recognition (MMR) systems provide a fully automatic framework to recognize and classify different vehicle models. Several approaches have been proposed to address this challenge, however they can perform in  restricted conditions. Here, we formulate the vehicle make and model recognition as a fine-grained classification problem and  propose a new configurable on-road vehicle make and model recognition framework. We benefit from the unsupervised feature learning methods and in more details we employ Locality-constraint Linear Coding (LLC) method as a fast feature encoder for encoding the input SIFT features.	The proposed method can perform in real environments of different conditions. This framework can recognize fifty models of vehicles and has an advantage to classify every other vehicle not belonging to one of the specified fifty classes as an unknown vehicle. The proposed MMR framework can be configured to become faster or more accurate based on the application domain.  The proposed approach is examined on two datasets including Iranian on-road vehicle dataset and  CompuCar dataset. The Iranian on-road vehicle dataset contains images of 50 models of vehicles captured in real situations by traffic cameras in different weather and lighting conditions.  Experimental results show superiority of the proposed framework over the state-of-the-art methods on Iranian on-road vehicle datatset and comparable results on CompuCar dataset with 97.5\% and 98.4\% accuracies, respectively.
	\end{abstract}
	
	
	\IEEEpeerreviewmaketitle

	\section{Introduction}
	\IEEEPARstart{F}{ine}-grained classification~\cite{IEEEhowto:Hillel}-\cite{IEEEhowto:Berg} as comes with its name, is a classification framework where the input data is assigned to very fine class labels. In this approach, the class labels are visually very similar, with very minor differences. Following several promising results~\cite{fg1,fg2,fg3}, fine-grained classification has recently attracted researchers to improve the classification accuracy one step further.

	Several problems~\cite{fgd1,fgd2,fgd3} have been targeted to be used as benchmarks for new proposed fine-grained approaches, which mostly are categorized into two main streams. In the first stream researchers have applied their proposed methods on standard dataset such as flower102~\cite{IEEEhowto:Flower102} or bird-200-2011~\cite{IEEEhowto:Bird}. On the other hand, there are several new frameworks proposed to address real-world problems such as face recognition and highway traffic data analysis. One of the real-world problems, which is considered as a fine-grained classification, is vehicle make and model recognition. This application is mostly utilized in intelligent transportation systems and frameworks. 
	      
	Intelligent transportation systems (ITS) have been growing fields of research for the past decade due the tremendous growth in the number of vehicles, and their crucial roles in the human life. Categorization of vehicles in roads is  the stepping-stone of several applications in ITS.  The vehicle detection and  classification, specially the vehicle Make and Model Recognition (MMR) system, which aims to recognize  different types and models of vehicles based on pre-defined categories, is becoming an important area of research in computer vision as well as in ITS. 
	
	The most common and the oldest  approach to address this problem is to capture the license plate of the vehicle with the help of a license plate recognition (LPR) system and to  identify the vehicle model by searching the license plate within available databases and then finding the make and model assigned to the license plate.  However, the first requirement to apply this approach is to access the national vehicle information databases. Furthermore, the LPR systems are not completely accurate and  their performance sometimes becomes erroneous. To improve the performance of LPR systems, fairly high resolution camera equipments are required to cope with several challenges  such as sun-light reflection, varying weather, and poor light conditions, which make the system expensive to operate. 
	
	The vehicle make and model recognition (MMR) system is a solution to the above problem, which improves the accuracy of LPR systems. The MMR not only helps the LPR systems to improve their performance, but also helps to find traffic violations such as license plate cloning and escaping from police.
	
	An MMR system consists of two main parts; detection of the vehicle in the input image, followed by identification of the make and model for the vehicle appeared in the sub-image containing the object.
	
	Here the vehicle make and model classification is addressed via a fine-grained classification framework. 	A unified MMR framework -- so-called \mbox{ORV-MMR} (i.e., On-Road Vehicle Make and Model Recognition) is proposed, where a vehicle is classified to one of the fine class labels. The proposed framework recognizes 50 different classes of Iranian on-road vehicles. The classes vary significantly, including  bus, mini-bus, truck, mini-truck, van, off-roader and sedan. The proposed framework can recognize the most common  Iranian on-road vehicles commuting in the domestic roads.  Furthermore, two extra classes are defined, such that the system classifies the unknown vehicles (the vehicles not belonging to any of the previously defined 50 classes) into these two extra classes and avoids misclassification. These two classes are named as ``Unknown Light'' and ``Unknown Heavy'' and  the vehicles  not belonging to any of the 50 predefined classes are to be classified as either ``Unknown Light'' or "Unknown Heavy''  based on their sizes (i.e., light or heavy), respectively. It is to notice that, the utilized dataset includes numerous challenging samples and the proposed framework is designed to address all challenges.   
	
	The proposed  ORV-MMR framework is designed to work in real-world situations, where there is no limit on lighting condition, field of view or the size of vehicle appears in the image. The proposed framework is examined on an Iranian on-road vehicle dataset whose images are captured are different weather conditions and with a wide range of variations. The images were extracted by speed-camera systems with various field of views, illumination, sun light reflection, and different weather conditions, which  make the recognition complicated since the detection of accurate vehicle ROI is difficult. Indeed the large variation in the vehicles size and body in this dataset makes the ROI detection a very challenging task. 
	
\begin{figure}[t]
		\centering
		\includegraphics[width=1\linewidth]{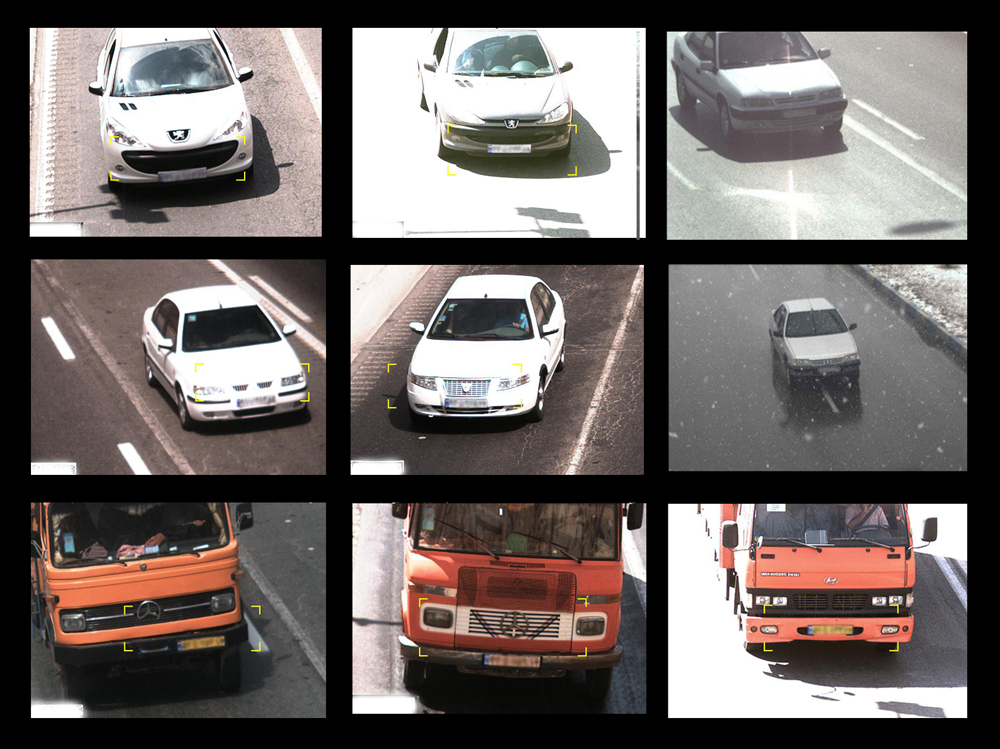}
		\caption{Iranian on-road vehicle  dataset examples. As seen, images were captured at different weather conditions, illuminations and filed of views. The sun light reflection in some situations saturates the images and makes some part of vehicle unrecognizable as shown in the second and third images in the first row. }
		\label{fig:image1}
	\end{figure}

	 Figure~\ref{fig:image1} shows some examples of Iranian on-road vehicles dataset at different conditions. Evidently the images were captured at different lighting conditions. The sun light reflection in some situations saturates the image intensities and as a result some parts of vehicle is not even visible in the image. In addition to the aforementioned limitations and constraints, the proposed framework must process the images in a real-time manner.
	 
	This paper provides a solution for the MMR problems when the part detection and ROI selection is not accurate and the input images have different illumination, scale and point of view conditions. In such scenarios, the proposed algorithm outperforms the standard convolutional neural network (CNN) methods. Furthermore, the contributions of this paper can be folded into several parts:
	\begin{itemize}
		
			\item We propose an unsupervised feature extraction  method to mitigate the issue of standard CNN features. This method  provides a trade-off between modeling accuracy and running-time performance for operating at different visual situations.
			\item A fine-grained classification approach  (ORV-MMR)  is introduced to classify vehicles into fine class labels. The proposed framework also calculates a meaningful confidence for each classification result.
			\item A new technique is proposed to identify the input images not belonging to any of the pre-defined class labels and classify them into one of the defined unknown classes based on their appearances. 
	\end{itemize} 
	The following section describes related works in more  detail. The proposed framework and the main contributions are explained in Sec. III. In Sec. IV the experimental  results will be discussed and finally the conclusions will be presented in Sec. V.

	\section{Related works}
	The most challenging part of a MMR problem is that the between-class variance for the predefined class labels is very small, which makes the classification a complicated task.  This type of problem has been emerged recently and is known as fine-grained classification challenge, where the class labels are distributed with very small between-class variances.
	
  A common approach to solve the fine-grained problem  is to find  discriminant parts, then extract features from the input images and combine the extracted features to build the final feature vector for classifying the input. Classification of birds to their specific species is an example of fine-grained classification which extracts features from discriminant parts and classifies them to finer class labels, which determines their species~\cite{IEEEhowto:Z.Ge,IEEEhowto:N.Zhang}. As such, vehicle make and model recognition (MMR) problem can be categorized as  a fine-grained classification problem when the number of class labels is increased in the modeling framework. Krause {\it et al.}~\cite{IEEEhowto:Krause-Learning} used convolutional neural networks (CNNs) to extract discriminative parts and features for fine-grained classification. 
  
  Fang {\it et al.}~\cite{IEEEhowto:Fang} proposed a coarse-to-fine CNN framework to detect the most discriminant parts of vehicle for feature extraction. They used a one-versus-all SVM classifier and evaluated their method using the CompuCar dataset~\cite{IEEEhowto:Yang-CompuCar}. Biglari{\it et al.}~\cite{IEEEhowto:CascadePartBased} proposed a cascade part-based approach for vehicle MMR problem. The authors used SVM for classification and CompuCar dataset for evaluation. 
  
  AbdelMaseeh {\it et al.}~\cite{IEEEhowto:AbdelMaseeh} introduced a method for extracting discriminative parts which are annotated by a human expert from the input images and using global shape and appearance descriptors~\cite{IEEEhowto:Belongie} for vehicle recognition. Psyllos~{\it et~al.} \cite{IEEEhowto:Psyllos} utilized a license plate detection technique to localize the plate location within the car frontal view images and then extracted the vehicle's logo. The SIFT features \cite{IEEEhowto:Lowe} were extracted from the cropped sub-image of the vehicle's logo for the purpose of classification, where a probabilistic neural network (PNN) was used to identify the class labels. Additionally, the authors used a color recognition method for recognizing vehicle's color. 
	
	Llorca {\it et al.}~\cite{IEEEhowto:Llorca} proposed a MMR system that models the geometry and appearance of vehicle from rear view images by the help of license plate recognition module. This approach was utilized in~\cite{IEEEhowto:Psyllos-2} where symmetric measurements were provided  to detect and to describe a vehicle.
	
	
	Several approaches  have been proposed to extract the robust features such as curvelet transform~\cite{IEEEhowto:Kazemi}, SIFT~\cite{IEEEhowto:Dlagnekov}, bag-of-words~\cite{IEEEhowto:Baran} and even CNN-based features~\cite{IEEEhowto:Yang-CompuCar} from the entire or a selected ROI of input image.
	The  SIFT features were successfully applied by Dlagnekov \cite{IEEEhowto:Dlagnekov}. Baran \mbox{\it et al.} \cite{IEEEhowto:Baran} proposed two different approaches for MMR problem to separately address running-time speed (i.e., real-time) and accuracy. They used a set of descriptors such as SIFT, SURF and MPEG-7~\cite{IEEEhowto:Manjunath}, which were fed into a classifier as the final step of that framework.
	
	Besides hand-crafted features like SIFT,  unsupervised feature learning techniques such as bag of words (BoW)~\cite{IEEEhowto:Csurka}  were vastly used to address the vehicle MMR problem. 
	Zafar~{\it et al.}~\cite{IEEEhowto:Zafar} proposed a 2D-LDA~\cite{IEEEhowto:Li} to extract features from the pre-defined ROI of vehicle images. Jang {\it et al.}~\cite{IEEEhowto:Jang} used the idea of BoW created based on the speeded up robust features (SURF)~\cite{IEEEhowto:Bay} with a structural verification technique. Their method was examined on a realistic-looking toy car dataset. Siddiqui {\it et~al.}~\cite{IEEEhowto:Siddiqui} applied SURF features and single and modular dictionary learning frameworks to address MMR problem.  Petrovic and Cootes~\cite{IEEEhowto:Petrovic} proposed a feature representations for rigid structure recognition to identify objects with an abundant number of categories.
	
	Beside extracting the robust features, selecting appropriate classifier for a MMR framework is another important challenge in vehicle MMR problem.  Neural Networks (NN)~\cite{IEEEhowto:Psyllos2008}, Support Vector Machine~\cite{IEEEhowto:Cortes}, Bayesian methods~\cite{IEEEhowto:Bernardo}, k-nearest neighbors~\cite{IEEEhowto:Cover} and na\"ive Bayes methods~\cite{IEEEhowto:Pearce} are the most particular classifiers utilized for this problem. Choosing an effective classifier to tackle such a problem is a challenging task since the between-class variances is very small while the within-class variance has opposite behavior and is very large.

	Most of current MMR frameworks were examined via benchmarks with limited imaging conditions, with vehicles of small size and scale variation, and within some specific field of views. Furthermore, the methods performance also depends on an accurately detected ROI~\cite{IEEEhowto:Pearce,IEEEhowto:Zafar}. In other words, in the MMR systems, the  input image is set to be the extracted vehicle's ROI without any clutter and background objects. These requirements reduce the usability and sustainability  of these types of system in real-world applications, where it is not possible to limit the imaging conditions, field of views or even the size of vehicles.

	\section{Methodology}
	
	In this section the proposed framework for classifying vehicles into  fine-grained class labels is explained. The proposed method is folded into three sections to provide a better explanation.  Next, we discuss the proposed unsupervised feature extraction for the purpose of fine-grained classification. Then, different parts of the vehicle make and model recognition systems are explained in more details and finally the proposed unknown class  identification will be proposed to identify those vehicle which are not belong to any pre-defined class label. 
	\subsection{Unsupervised Feature Learning}
	 \label{sec:UFL}
	 \sethlcolor{green}
	Unsupervised feature learning methods provide end-to-end approaches which can automatically extract the most useful features from input data.
	Although Deep neural networks and particularly convolutional neural networks (CNNs) benefit from the same issues but in some situations they still need some progresses. For instance, these types of methods can not work perfectly when there is not enough training data. Furthermore, the CNN approaches are very sensitive when they are being trained by an  unbalance dataset with a wide variety and different conditions. Among many unsupervised feature learning methods, we propose a new unsupervised feature extraction framework which employs some interesting modules to boost the performance.

	 The proposed feature extraction method is based on the learned bases, dictionary or codebook~\cite{IEEEhowto:Coates} approach where the framework is trained in an unsupervised manner. Here we take advantage of  dense-SIFT feature descriptor~\cite{IEEEhowto:Bosch} to address the huge variation of object appearance among training sample.  The SIFT feature descriptor and specially dense-SIFT feature descriptor have been demonstrated to be powerful in different situations  such as objects with different scales, different field of views and varied illumination conditions. Therefore, we utilize SIFT features  as the backbone of the proposed framework.
	 
	 
	 As the first step in this learning procedure, a sub-set of training data are selected randomly and a set of  features are extracted from the samples in the selected subset. After extracting the dense-SIFT features from the subset of training data, a set of features (i.e., approximately one million features) are collected to build the codebook. To learn the bases,  a k-means clustering method is employed and is performed on the collected dense-SIFT features which the center of each cluster is considered and stored as a basis. The number of clusters (bases) is set by $M$ which is a user-specified  parameter of the proposed method and can be determined based on cross-validation techniques. However it has been demonstrated ~\cite{IEEEhowto:Huang} that while increasing the number of bases ($M$) improves the classification accuracy, it also increases the training and testing run-time. Therefore, in applications with a real-time constraint, increasing the size of codebook is restricted by the running time criteria. Next, features are transformed into the new learned space.  
	 
	 \subsubsection{{Feature Encoding \& Transformation}}
	 Every input feature (extracted from an input image) is encoded and transformed into  the new learned space  by used of the trained  dictionary explained in the previous Section. In this step, each input feature descriptor triggers a number of bases, and creates a coding vector with the length equal to the number of bases. Several encoding schemes have been proposed by Huang {\it et al.}~\cite{IEEEhowto:Huang} including voting based, fisher vector, reconstruction based, local tangent and saliency coding methods. However the reconstruction-based coding approaches and local tangent approaches are the most appropriate approaches in term of adaptiveness and accuracy. One of the main advantageous of them is that they  can be computed independently when utilized in conjugation with other processes which is important in real-time applications. Here the reconstruction based coding technique is applied  for the purpose of real-time performance.
	 
	 To encode a new feature set $\rm x$ using the reconstruction-based approach, the objective function~\eqref{eq:objfun} is minimized to achieve the coefficients $\alpha$ given the input $\rm x$:
	 \begin{align}
	 \label{eq:objfun}
	 {{\mathop{\min }_{\alpha } \frac{{\rm 1}}{{\rm 2}}\|{\rm x-B}\alpha\| }}^{{\rm 2}}_{{\rm 2}}{\rm +}\beta \left(\alpha \right)
	 \end{align}
	 where  $B$ is a set of the bases, $\alpha$ is the coefficient set that needs to be optimized and $\beta(\cdot) $ is a regularizer function. Different regularizer functions, generate different reconstruction based encoding methods. Nazemi {\it et al.}~\cite{IEEEhowto:Nazemi} evaluated a simple and fast sparse feature coding method on a limited MMR dataset to address the  running time issue while it preserves the modeling accuracy. Here we take advantage of  a more complex reconstruction based method (locality-constraint linear coding (LLC))~\cite{IEEEhowto:Wang} which provides a fast   implementation of local coordinate coding~(LCC)~\cite{IEEEhowto:Yu-2009} and utilizes the locality constraint to project each descriptor into its local coordinate (basis) space. As such,~\eqref{eq:objfun} can be reformulated as:
	 \begin{align}
	 \label{eq:lcc}
	 \mathop{\min }_{\alpha }\left( {{\frac{{\rm 1}}{{\rm 2}}\|{\rm x-B}\alpha\| }}^{{\rm 2}}_{{\rm 2}}{\rm +}\sum^{{\rm M}}_{{\rm i=1}}{{{\rm (}\alpha \left({\rm i}\right){\rm exp(}{\left\|{\rm x-}{{\rm b}}_{{\rm i}}\right\|}_{{\rm 2}}{\rm )/}\sigma {\rm )}}^{{\rm 2}}}\right) 
	 \end{align}
	 \noindent where $b_{{\rm i}} \in B$ is the ${{\rm i}}^{{\rm th}}$  basis of the dictionary $B$, ${\rm M} = |B|$ is the number of bases in $B$ and $\sigma $ controls the size of locality span.
	 
	 To decrease the computational complexity of~\eqref{eq:lcc} and encode a new feature faster, ${\rm K} \leq {\rm M}$ nearest bases of the $B$ to input $x$ are chosen and  a new dictionary  $\tilde{B}$ is formed which~\eqref{eq:lcc} is reformulated via $\tilde{B}$:
	 \begin{align}
	 \mathop{\min }_{\alpha }\left( {{\frac{{\rm 1}}{{\rm 2}}\|{\rm x- \tilde{B}}\alpha\| }}^{{\rm 2}}_{{\rm 2}}{\rm +}\sum^{{\rm K}}_{{\rm j=1}}{{{\rm (}\alpha \left({\rm j}\right){\rm exp(}{\left\|{\rm x-}{{\rm b}}_{{\rm j}}\right\|}_{{\rm 2}}{\rm )/}\sigma {\rm )}}^{{\rm 2}}}\right) 
	 \label{eq:llc1}
	 \end{align}
	 where $b_{{\rm j}} \in \tilde{B}$. 
	 Finding the ${\rm K}$ nearest neighbors is a time consuming step in the reconstruction-based encoding technique. To further speed up this process, a kd-tree structure~\cite{IEEEhowto:Muja} is used to find the $K$-nearest neighbors. In a kd-tree structure, it is possible to determine the maximum number of comparisons to find the \mbox{$K$-nearest} neighbors which highly decreases the computational complexity.
	 
	 
	 \begin{figure*}[t]
	 	\centering
	 	\includegraphics[width=1\linewidth]{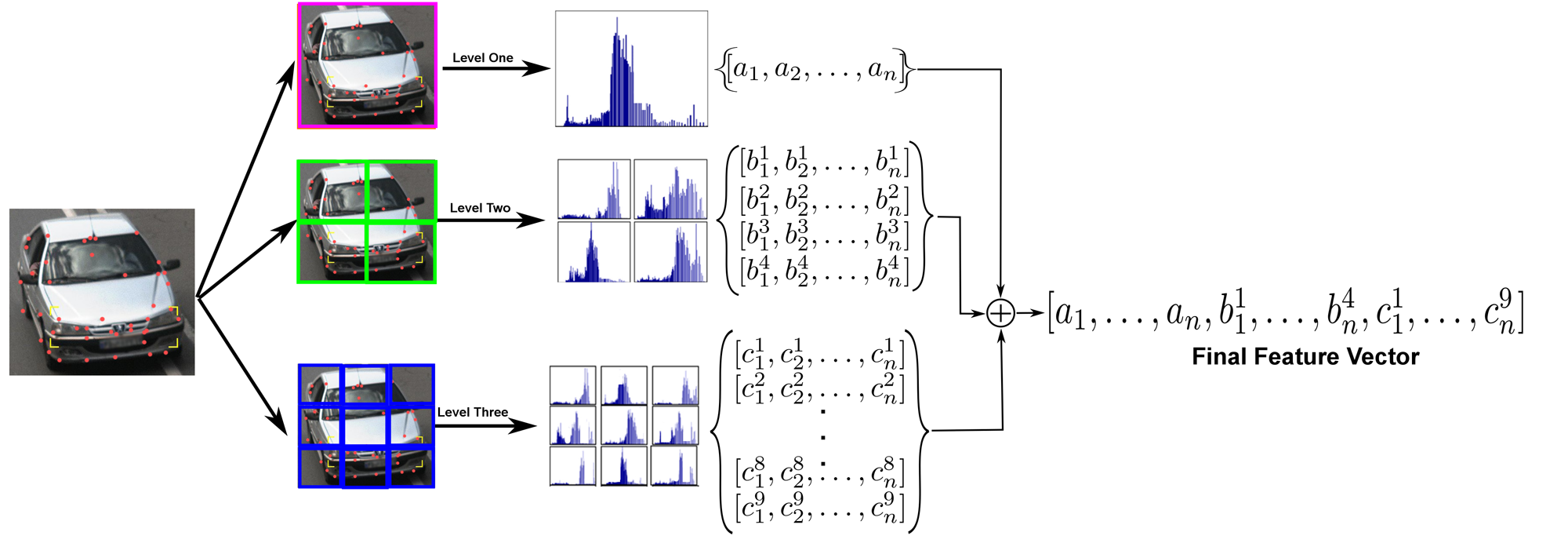} 
	 	\caption{A Spatial pyramid matching (SPM) method that is applied on the  features extracted from the input image in three levels with 1, 4 and 9 grids respectively. In level one all Sift features are pooled (merged) together and results a single feature vector. However in  level two and three, features located in each  grid are pooled together and results 4 and 9 individual feature vectors. At the end all  14 ($1+4+9$) feature vectors are concatenated to build the final feature vector.}
	 	\label{fig:SPM}
	 \end{figure*}

	 \begin{figure*}[t]
	 	\centering
	 	\includegraphics[width=1\linewidth]{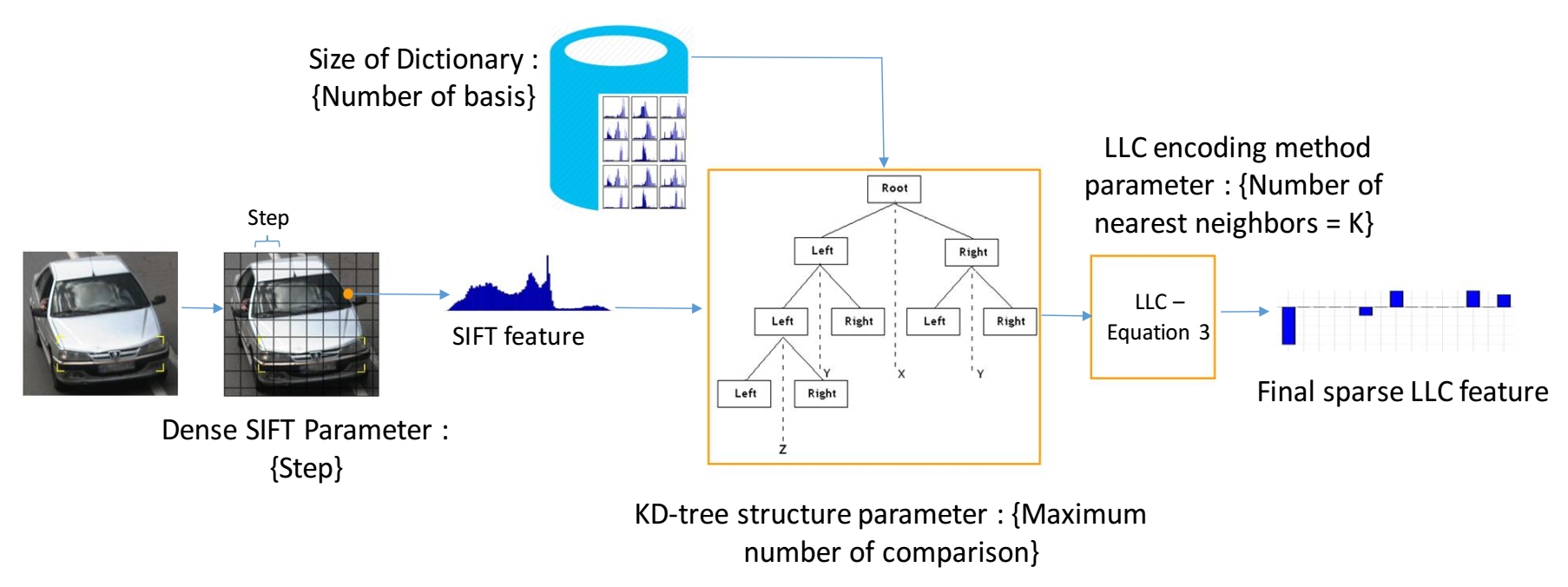}
	 	\caption{This picture shows the feature learning pipeline and its important parameters.This pipeline starts with one single SIFT descriptor and ends with a single sparse LLC feature. The first parameter is Step which defines the number and the position of SIFT features in the input image. Second parameter is the size of dictionary. Next two parameters are Maximum number of comparison of Kd-tree and number of nearest neighbors in LLC algorithm which affect the time and accuracy of LLC encoding.}
	 	\label{fig:Feature}
	 \end{figure*}

	 \subsubsection{Spatial pooling}
	 One of the key aspect of convolutional neural network is the extraction of spatially invariant features provided by use of pooling layers in network architectures. Here we take advantage of pooling by applying  the spatial pyramid matching (SPM) \cite{IEEEhowto:Lazebnik} approach. SPM  is performed  by partitioning the image into increasingly fine sub-regions.  After partitioning of the image,  features in the same  spatial grid are merged together.  SPM helps the feature encoding methods to preserve the spatial information of the input image and also  makes the features robust against the translation and reduces the dimension of the final feature vector as well. As seen in Figure~\ref{fig:SPM}, the features are pooled in a multi-scale griding scheme and at the end all pooled features are concatenated together to form the final feature vector. There are many pooling methods such as max, average, sum, log-mixture and weighted sum pooling. In this framework the max pooling approach is utilized combined with the LLC method since max pooling method is the best pooling method for merging the sparse features and keep the sparsity of the final feature vector~\cite{IEEEhowto:Boureau}.
	 \subsubsection{Configurable parameters}
	 Although, the proposed method of section A is a classical method for image classification, finding the best configurable method which can alternatively performs real-time and accurate even more than standard CNN features is the first contribution of this paper. The step and size parameters of Dense-SIFT, the number of bases, K nearest neighbors of LLC and the maximum number of comparison of kd-tree structure can be tuned and as a result improves the performance of proposed framework. Figure \ref{fig:Feature} shows the main parameters of the proposed unsupervised feature learning and where they are applied in the framework.

	The proposed unsupervised feature extraction framework is utilized to design a complete system for vehicle make and model recognition (ORV-MMR). The system has the capability of recognizing vehicle in real-world scenario and covers a wide range vehicles with fine class labels.   
	In the next section, the proposed ORV-MMR framework is explained in more details. 
	\subsection{On-Road Vehicle Make \& Model Recognition}
\begin{figure*}[!t]
		\centering
		\includegraphics[width=0.7\linewidth]{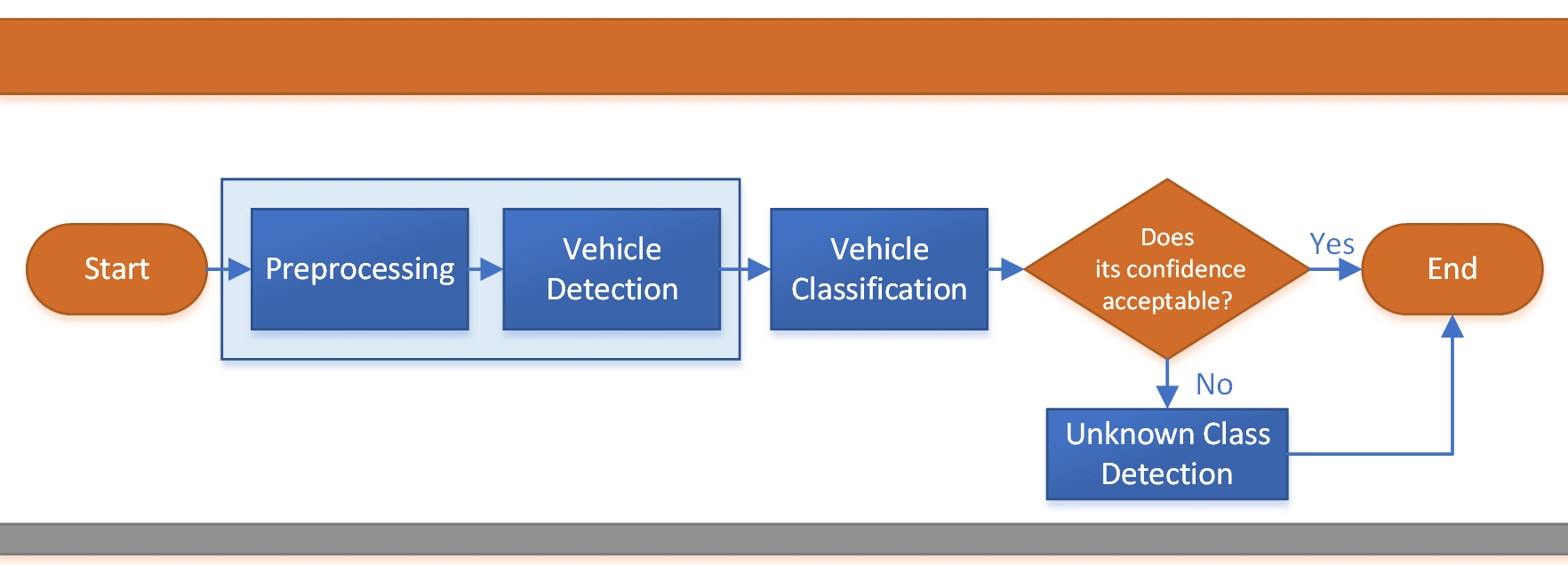}
		\caption{The flow-diagram of the proposed ORV-MMR system. The system is divided into three main parts including the ``preprocessing and vehicle detection", ``vehicle classification" and ``unknown classes detection".}
		\label{fig:Microsoft_Visio_Drawing1}
	\end{figure*}

	The vehicle make and  model classification  is formulated as a fine-grained image classification problem where the extracted features via the proposed unsupervised feature extraction framework is utilized to discriminate different vehicles.
	Figure~\ref{fig:Microsoft_Visio_Drawing1} demonstrates the flow-diagram of the proposed ORV-MMR system. As seen, a comprehensive ORV-MMR framework classifies an image in three steps. In the the first step, the image is preprocessed and the vehicle is detected in the scene, then the features are extracted and the vehicle is classified into one of 50 pre-defined class labels and at the end, based on the confidence of selected class, it is determined whether the vehicle is belonged to pre-defined classes or it should be assigned to one of the unknown classes.

	\subsubsection{{Preprocessing and vehicle detection}}
	The first part of the proposed ORV-MMR system is to detect the vehicle in an image and extract  the vehicle's bounding box. Varied illumination conditions is an issue to find the bounding box of the vehicle in the image.  Therefore, the contrast limited adaptive histogram equalization technique (CLAHE)~\cite{IEEEhowto:Zuiderveld} is applied to eliminate the illumination variations and the median filter is utilized to reduce the nonlinear noises which are produced by digital camera's sensor. This phenomenon usually occurs when an image is taken  in a dark environment.
	
\begin{figure}[t]
		\centering
		\includegraphics[width=1\linewidth]{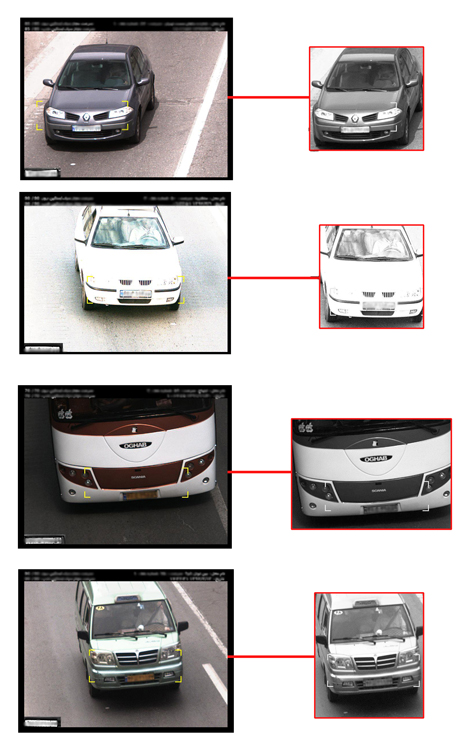}
		\caption{Some examples of the result of detection step of the proposed system. As seen, left column shows the input images while right column demonstrates the detected vehicles images.}
		\label{fig:Detection2}
	\end{figure}

	Finding the best ROI corresponding to the vehicle is another challenging problem and the extracted ROI is usually  inaccurate due to different lighting conditions. Inaccurate ROIs can fool the classifier as the vehicle's  shape contains discriminative information and an inaccurate ROI  poses a serious impact on the modeling accuracy. To address this issue and to make the detection part more accurate, a deformable part model (i.e., latent SVM)~\cite{IEEEhowto:Felzenszwalb}  was trained and utilized to detect the entire vehicle in the image. As a state-of-the-art method, latent SVM is based on the boosted cascades of classifiers for detecting the objects in the image~\cite{IEEEhowto:Girshick-lsvm} 
	and is one of the most applicable solution for real-world vehicle detection problems~\cite{IEEEhowto:Detection_Survey}. \mbox{Figure~\ref{fig:Detection2}} demonstrates some example results of performing the latent SVM approach on input vehicle images. As seen, the latent SVM method can extract vehicles from the image in different illumination variations.
	
	\subsubsection{{Vehicle Classification}}
	The ultimate goal of a MMR framework is to recognize the class of vehicle. However this step is the bottleneck of a MMR framework as  the  features should be  extracted and fed into the classifier; therefore, designing an accurate feature extraction approach has crucial impact on  the running time and the accuracy of the framework. 
	
	 We take advantage of feature learning within the proposed unsupervised framework explained in Section~\ref{sec:UFL}. The extracted features are transformed into another domain via the learned bases  and then based on a multi-scale approach they are pooled to result the final set of features. The whole process is performed in one computational layer  which makes it fast to evaluate compared to CNNs methods comprising of several processing layers. Since the framework is designed  within an unsupervised approach, it also relaxes the demand for a large training dataset and also it can address the unbalanced issue of the available dataset.
	
A final feature vector for each input image is achieved after performing the spatial pyramid matching. The  dimension  of the final feature vector is \mbox{$M\times \left(S_1+S_2+\dots +S_l\right)$} where $M$ is the number of bases and $S_i$ is the number of grids in $ i^ {th} $ level of spatial pyramid. The proposed framework uses a linear SVM for classification in the first stage and then a multi-layer perceptron neural network to specify the confidence of classification in the second stage.

It is worth to note that utilizing a nonlinear SVM classifier for large image classification problems is not recommended. The complexity of nonlinear SVM is $O(n^2\ \sim \ n^3)$ in the training stage and $O(n)$ in the testing one, where $n$ is the training size, while using a linear SVM the complexity is reduced to $O(n)$ in the training and a constant in the testing stage.  
%
	
\subsubsection {Classification Confidence}
\label{sec:MSC}
Providing  confidence level for the classification results is an important aspect of a system. As mentioned, SVM is applied for the classification purposes in this framework. However, SVM methods cannot produce any confidence level for the classification results. Here a new approach is applied to produce the classification confidence. Here, we take advantage of a multi-stage classification approach where the results of SVM classifier are passed to another classifier to determine the confidence of the classification.In other words, multistage classification structure contains two level of classifiers. In the first stage a bunch of SVM classifiers are used for classification and in the second stage, a multi-layer perceptron ($MLP$) neural network is used to produce the confidence for classification.

\begin{figure*}[!t]
	\centering
	\includegraphics[width=1\linewidth]{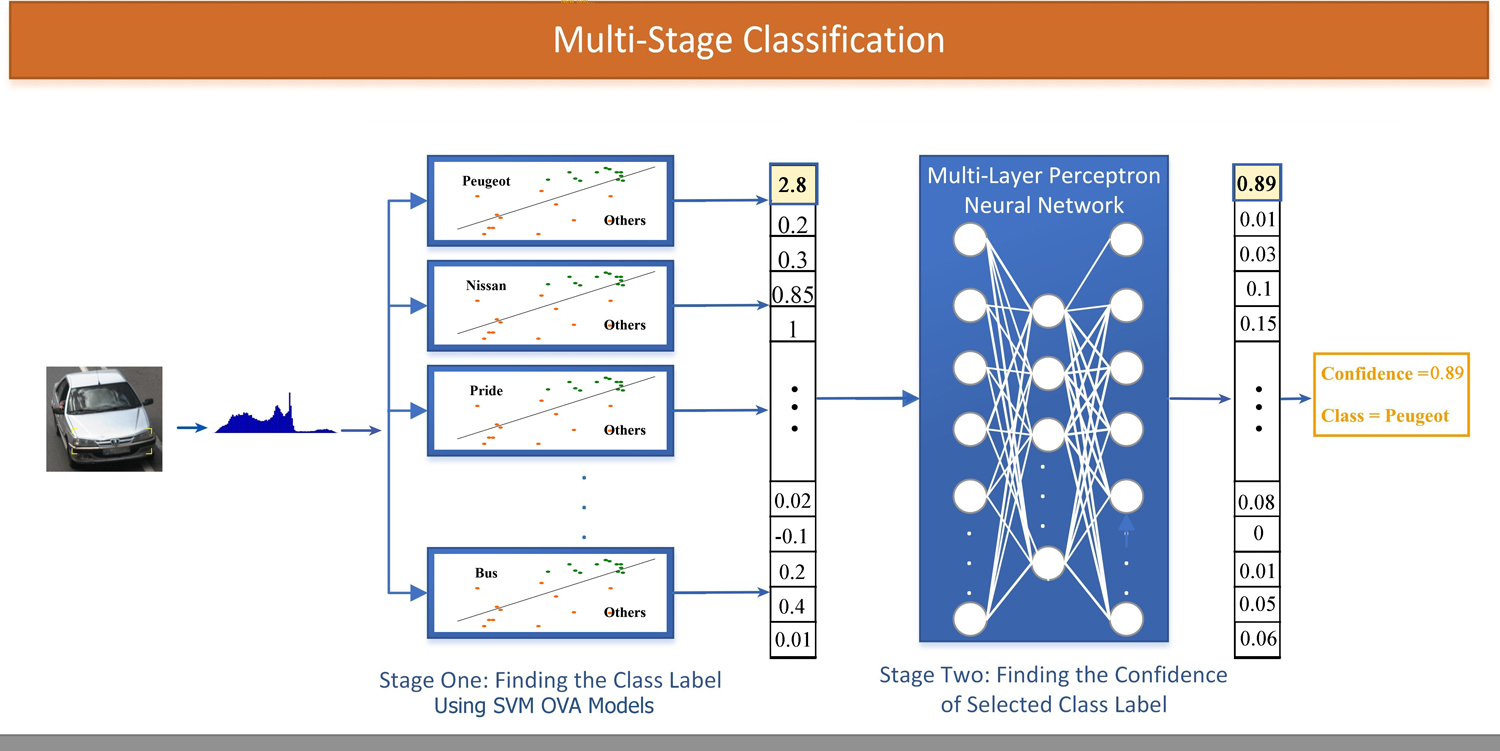}
	\caption{The structure of the multi-stage classification method which contains one-versus-all SVM models and a multi-layer perceptron neural network. The first stage contains one versus all (SVM) models and the output is a 50 dimensional vector. The second stage maps the 50-dimensional vector of the SVM's score into the class labels and provides the confidences. Here the first stage finds the class label which is Peugeot and the confidence is provided by the second stage.}
	\label{fig:Microsoft_Visio_Drawing6}
\end{figure*}

Figure \ref{fig:Microsoft_Visio_Drawing6} shows the structure of the  proposed multi-stage classification method to compute the classification confidence. 
A one-versus-all technique is utilized for training the SVM models. After training the SVMs (50 SVM models),  a $MLP$ is utilized to learn the final class labels' confidence in the second stage of classification. The inputs of this $MLP$ are the vectors consist of the output scores of 50 trained SVM models which are employed on training data and the outputs of this $MLP$ are the binary vectors consist of training labels.
It should be noted that the decision making and classification is done by finding the maximum score of trained SVM models' output at first stage while the second stage is performed to evaluate the confidence of the classification results.

\subsection{Unknown Class Identification}
A desirable MMR system should be able to identify all type of vehicles; however, creating such a system needs tremendous amount of training data. Furthermore, increasing the number of class labels in a classifier attenuates the performance of the system and decreases the overall modeling accuracy of the MMR framework. One solution to this problem is to find the specific number of class labels which covers the highest percentage of existent vehicle make and models. Here 50 most common  vehicles in Iran were selected as pre-defined class labels. However, to cover the whole variation of  vehicles and be able to identify  vehicles which are not in the pre-defined 50 class labels, a new class label is designed and  named as {\it unknowns}. A training dataset was collected for this specific  class label comprising  a set of vehicles  that  are not belonged to any of the pre-defined classes. Since the within-class variance of this category is very high,  the unknown class is divided into two sub-classes including the {\it unknown  heavy} class and {\it unknown  light} class. 

Due to the large variation of vehicles belong to this unknown class (i..e, unknown), considering these vehicles as two new classes and train the classifier with 52 class labels equally  is not feasible and it would reduce the classification accuracy. To address this issue,  the classification of these types of images is formulated as an anomaly detection approach. To the best of our knowledge, this is the first time that the object classification is addressed by an anomaly detection framework and it is another novelty of the proposed framework.

There are several techniques for anomaly detection. Chandola {\it et al.}~\cite{IEEEhowto:Chandola} categorized  anomaly detection techniques into five categories including  classification based, clustering based, nearest neighbor based, statistical, information theoretic and spectral techniques. Here a classification based approach is utilized to address two novel class labels.

\begin{figure*}[t]
	\centering
	\includegraphics[width=1\linewidth]{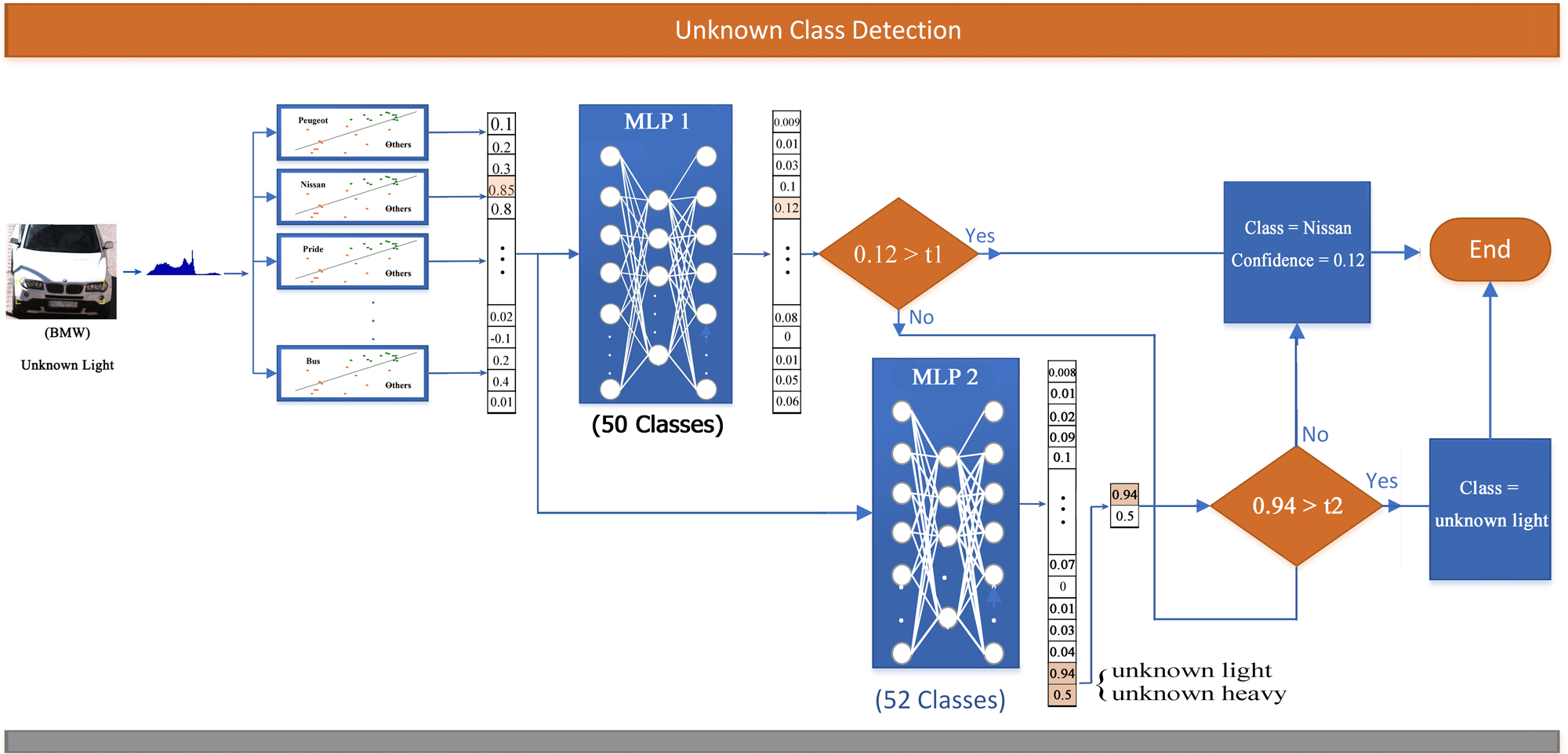}
	\caption{The flow diagram of two unknown classes detection framework. As seen, the input of the second MLP is the output of the first stage of multi-stage classification (MSC) model. If the confidence of the first NN is bigger that $t1$ then the input would be labeled as one of 50 known classes and if it is not, the second threshold decides if input is an unknown vehicle or not. In this figure, the input is a BMW car which is not in our 50 classes and so it is an unknown light vehicle. The first MLP says that the input is Nissan with the confidence of 0.12. Since the confidence is not bigger than predefined threshold t1 (t1 is 0.87) we refers to second MLP which is trained on 52 classes and includes unknown classes. The new confidence is bigger than t2 (t2 is 0.93) and we assigned the input into the unknown light class.}
	\label{fig:Microsoft_Visio_Drawing8}
\end{figure*}

For unknown class identification, we modify the  multi-stage classification structure which is illustrated in figure~\ref{fig:Microsoft_Visio_Drawing6} and add another multi-layer perceptron NN and two thresholds for detecting the unknown classes.
As shown in Figure~\ref{fig:Microsoft_Visio_Drawing8}, a multi-layer perceptron ($MLP 2$) neural network  is used to detect this behavioral anomalies (i.e., recognizing two unknown vehicle classes).  The scores resulted from 50 trained SVMs are utilized  as the input for a new MLP structure. The $MLP 2$ is trained to classify the input features (i.e, 50 SVMs scores) into one of 52 class labels (50 pre-defined class labels and two unknown class labels). Additionally, to make the identification process more robust, a two-level thresholding approach is utilized.

The two-level thresholding approach is utilized  to make the classification model more accurate. It is obvious that the classification accuracy for 50 class labels decreases when we add two more unknown classes, therefore, to decrease the effect of the new classification step, we take advantage of a confidence thresholding procedure. As explained before, the input image is first classified into one of 50 pre-defined class labels and then  the 50 confidence values of second classification stage ($MLP 1$) is utilized to determine whether the input image is one of the 50 pre-defined class labels or it belongs to one of the unknown classes. As shown in Figure~\ref{fig:Microsoft_Visio_Drawing8} for an input unknown light image, first the system compares the classification output confidence of the input sample (i.e., the confidence of predicted class label) with ${t}_1$, and  if the confidence is greater than ${t}_1$ then it can be labeled as one of 50 class vehicles but if the confidence is smaller than ${t}_1$, the system uses the $MLP 2$ model to determine whether the input image  belongs to  one of the unknown classes. If the  confidence of second $MLP$ model is greater than ${t}_2$ and the vehicle  belongs to one of the unknown classes then the vehicle is labeled as one of two unknown classes. Otherwise, if it does not belong to one of the unknown classes or its confident is less than even ${t}_2$, system trusts the first provided classification label.

\section{ Results \& Discussion }
The proposed ORV-MMR framework is  examined by the Iranian on-road vehicle dataset comprising a wide range of vehicle images captured in varied field of views, weather conditions and illuminations. The performance of the proposed framework and the competing methods are also evaluated by a second dataset  which is introduced in~\cite{IEEEhowto:Yang-CompuCar}. Since the competing methods are evaluated only based on the classification accuracy and it is assumed  that the vehicle's region of interest is extracted, the methods are examined only based on a cropped image,  only the classification stage  of the proposed ORV-MMR framework is compared with the state-of-the-art methods subject to the accuracy and their running time. To have a comprehensive evaluation, the efficiency and efficacy of the proposed framework is also measured  considering the combination of vehicle detection and vehicle classification performance based on the  Iranian on-road vehicle dataset. Additionally, the effects of framework's parameters which mentioned in section III.A are evaluated.

\begin{figure*}[!t]
\centering
\includegraphics[width=1\linewidth]{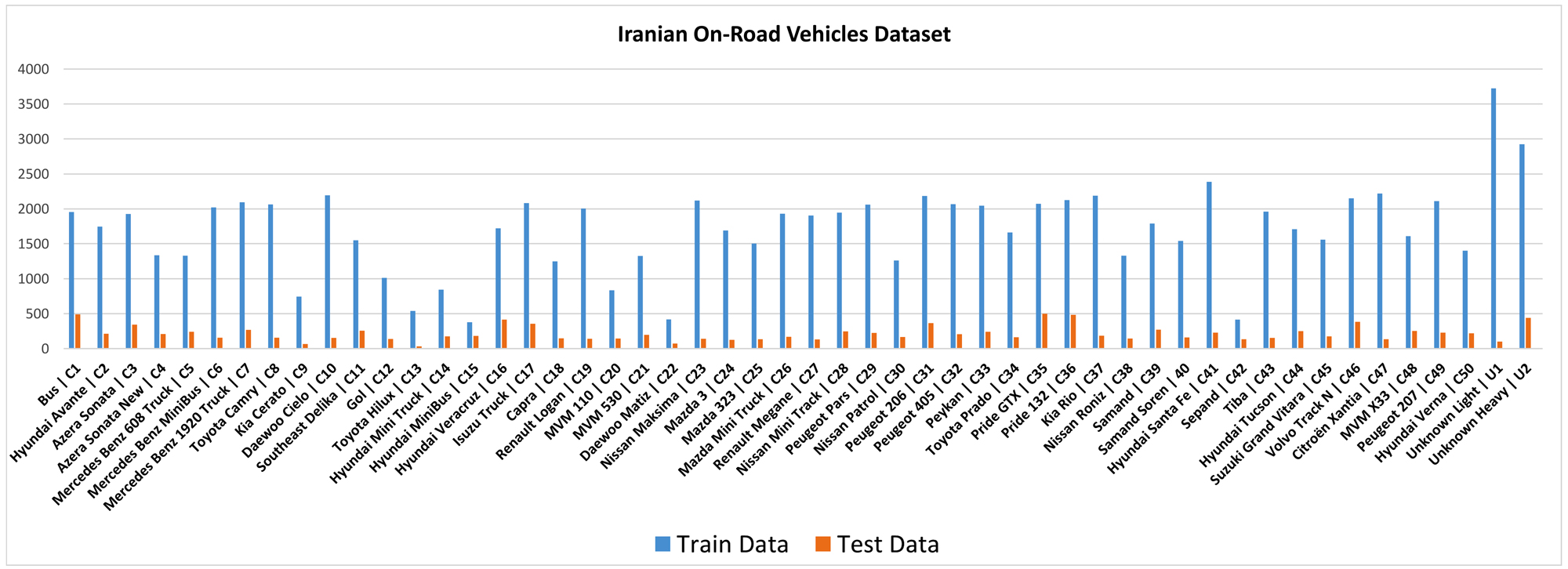}
\caption{The distribution of training and testing set of Iranian on-road vehicle dataset; The chart shows the number of training and testing samples for each class. The maximum training samples of the known 50 classes is around the 2000. The maximum number of testing samples is almost 500 samples.}
\label{fig:Data_52}
\end{figure*}

\subsection{Datasets}
The proposed ORV-MMR is evaluated by two different datasets:\\
I) \textbf{Iranian on-road vehicle dataset}, contains 93008 images of the most common Iranian on road vehicles captured in different scales, weather conditions, illuminations, and varied field of views by different cameras. Vehicle images were captured in a wide range of field of views by several cameras with different backgrounds. 81539 images from 50 known vehicle classes and 1117 images of two unknown classes (i.e., vehicles which are not belonged to one of those 50 known classes) were used as training set while 10352 images of 52 vehicles classes were utilized in the test set. The number of training samples per class labels are different for each class, however the maximum number of images per class label is about 2000 images in the training set. Figure~\ref{fig:Data_52} demonstrates the distribution of training and testing samples corresponding to each vehicle category and class. As seen, although we limited the number of training data per each class, the number of image samples per class are varied and in some categories the number of training images are much less than 2000 images evident by classes $\{C13, C15, C22, C42\}$ with less than 500 images per class label.

  Figure~\ref{fig:Objs} shows the representative example of each class label in the Iranian on-road vehicle dataset. Each vehicle category is  encoded by  $Cx$ where $x={1:50}$ for easier representation purposes. It is obvious that some class labels are very similar. For example, the vehicles with class labels $\{C29, C32, C47\}$ (green border) are very similar in terms of their structures and shapes. Some other examples are also demonstrated in Figure~\ref{fig:Objs} where the similar categories are highlighted by the same color border. As seen in Figure~\ref{fig:Objs}, the dataset contains various car types such as bus, mini-bus, track, mini-track, van, off-roader and sedan.
  
\begin{figure}[t]
	\centering
	\includegraphics[width=1\linewidth]{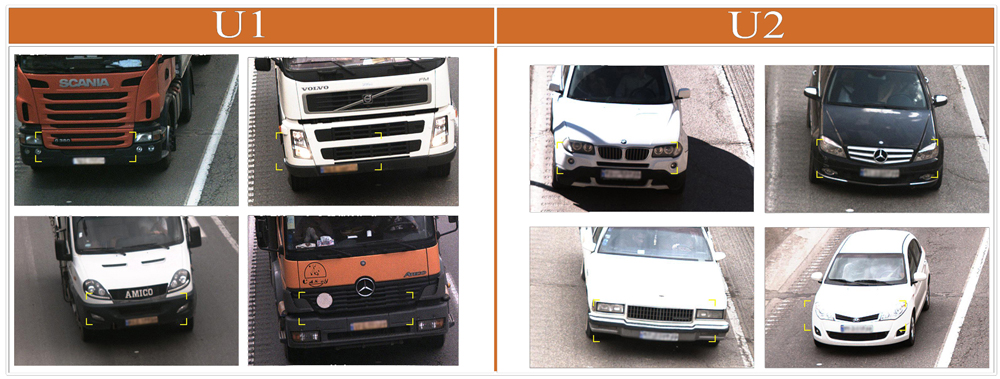}
	\caption{Some examples of two unknown classes; the Left set U1, shows some samples from the heavy unknown class while the right set, U2, indicates the light unknown class examples. It is worth to note, these two classes encodes several number of vehicle with different make and models which here just four samples are demonstrated for illustrative purposes. }
	\label{fig:Unknowns}
\end{figure}

 Another challenging aspect  of the Iranian on-road vehicle dataset is the unknown classes. As mentioned before, the proposed framework must classify the vehicles which are not belonged to the 50-classes, into one of the two-unknown classes. Figure~\ref{fig:Unknowns} shows some examples of two unknown classes. As seen, while the unknown vehicles  and the vehicles  of 50 known classes are very similar, they must be classified as unknown vehicle. Each unknown class label contains several vehicle make and models compared to the 50 known classes which are  represented only  by one specific vehicle make and model.

II) \textbf{CompuCar dataset~\cite{IEEEhowto:Yang-CompuCar}},   contains 44481 frontal images of 281 different models of cars. The vehicles are mainly  sedans and off-roaders. The dataset does not contain any type of heavy cars such as buses or tracks. 31148 images were used as training set and the remaining images were utilized in the testing stage. The number of vehicles per class in training set are between 14 to 565 vehicles and the average number is about 100 vehicles. It is worth to note that  the ROI of the vehicle in each image is provided (the vehicle is detected) and the vehicle images are aligned.
\subsection{Competing Methods}
The proposed framework is compared with two different state-of-the-art methods:
\begin{itemize}
	\item \textbf{AlexNet}: The AlexNet~\cite{IEEEhowto:AlexNet} network structure is employed as the first competing methods. AlexNet is one of the well-known deep neural network architecture  outperformed other methods in Imagenet classification challenge in 2012 and it is usually utilized for classification purpose. Here we compare the proposed  framework with  the AlexNet architecture as baseline for deep neural network solutions for the vehicle make and model classification problem.  AlexNet has five convolutional layers and two fully connected layers. The pre-trained network model on  Imagenet is utilized as the initialization and the network is fine-tuned for each datasets.  
	\item \textbf{F-G-VMMR}: Jie Fang {\it et al.}~\cite{IEEEhowto:Fang} proposed a coarse-to-fine CNN framework for detecting the most discriminant part of vehicle and extracting the features from detected parts.  We compare the proposed ORV-MMR framework with F-G-VMMR method which is considered as a complex deep neural network approach for the purpose of vehicle make and model classification.  Jie Fang {\it et al.}~\cite{IEEEhowto:Fang} used a one-versus-all SVM classifier for classification. Although their method works well on CompuCar dataset, results showed that it can not outperform the proposed framework on the challenging images of Iranian on-road vehicle dataset. 
\end{itemize}

\subsection{ORV-MMR Configuration Setup }
 The parameters $M$ (the number of basis in the dictionary) and  $K$ (the number of neighbors in the LLC methods) were set to \{1200,~3600\} and 5 respectively.  $\{1, 4, 9\}$ region parts \mbox{(1, 2 and 3 parts on each axis as shown in Figure~\ref{fig:SPM})} were considered as the parameters for the spatial pyramid matching while the max pooling is performed as the pooling scheme. The SVM and neural network are implemented by use of VLFEAT library~\cite{IEEEhowto:VLFEAT} and  Matlab environment. An extra validation set of unknown classes, which contains 1000 images for two unknown classes is used to tune two threshold parameters for recognizing the unknown classes (${t}_1,{t}_2$). The optimized threshold for ${t}_1$ is  0.87 while for ${t}_2$ is 0.93. To speed up the proposed algorithm, a kd-tree structure is used to find the nearest neighbors in the dictionary. A kd-tree is a data structure which is used to quickly solve nearest-neighbor queries. The kd-tree structure also facilitates to specify the  maximum number of comparisons per query and calculates approximate nearest neighbors. This parameter plays a key role in implementation of the proposed framework which affects the running time of the proposed method. The framework was configured by setting the maximum number of comparisons to $100$. The main advantages of dense-SIFT over SIFT is its computational complexity and running-time speed. We use the PHOW~\cite{IEEEhowto:PHOW} descriptors, a variant of dense-SIFT descriptors to speed up the computation of feature extraction. Two parameters of step and size of PHOW algorithm is set to 5 and $\{4,6\}$ respectively. 

 The MatConvNet~\cite{IEEEhowto:Matconvnet} toolbox is employed for training the CNN models. Training is performed on a personal computer with an NVIDIA GTX 980 GPU with 4GB memory. The VLFEAT open source library was  applied to perform the PHOW and the K-means algorithms.
\begin{figure*}
\vspace{- 0.3 cm}
\fboxsep=1mm
\fboxrule=2pt
\begin{center}
\begin{tabular}{cccccccc}
\includegraphics[width = 1.6cm]{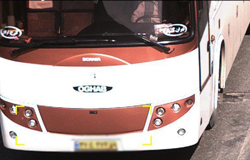}&
\includegraphics[width = 1.6cm]{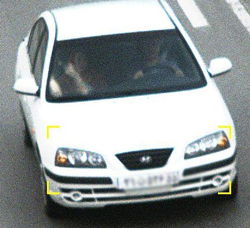}&
\includegraphics[width = 1.6cm]{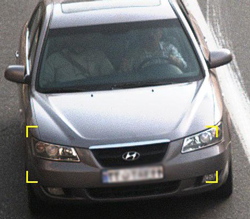}&
\includegraphics[width = 1.6cm]{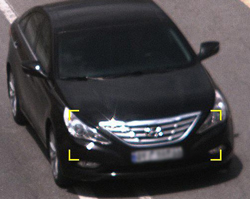}  &
\fcolorbox{red}{white}{\includegraphics[width = 1.6cm]{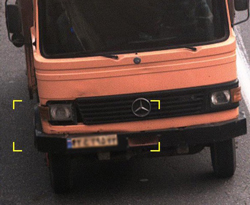}}&
 \fcolorbox{red}{white}{\includegraphics[width = 1.6cm]{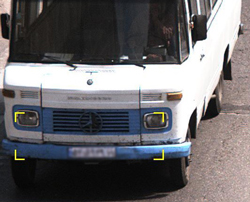}}  &
   \includegraphics[width = 1.6cm]{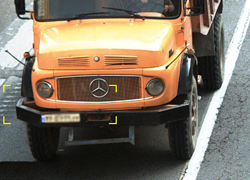}&
   \includegraphics[width = 1.6cm]{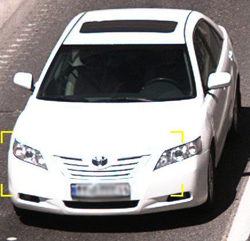}\\
      C1&C2&C3&C4&C5&C6&C7&C8\vspace{-0cm}\\
     \includegraphics[width = 1.6cm]{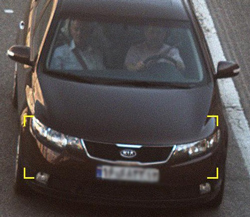}&
   \includegraphics[width = 1.6cm]{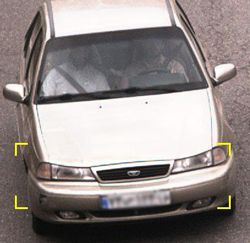}&
\includegraphics[width = 1.6cm]{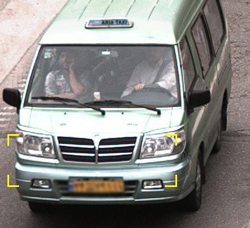}&
\includegraphics[width = 1.6cm]{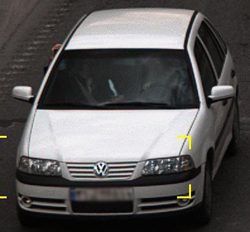}&
\includegraphics[width = 1.6cm]{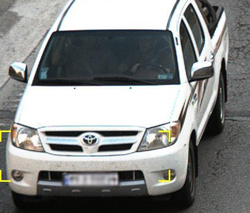}&
\includegraphics[width = 1.6cm]{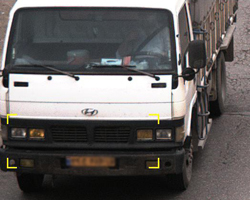}  &
\includegraphics[width = 1.6cm]{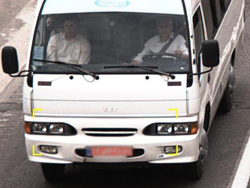}&
 \includegraphics[width = 1.6cm]{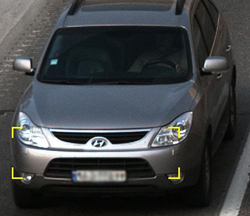}  \\
  C9&C10&C11&C12&C13&C14&C15&C16\vspace{-0cm}\\
   \includegraphics[width = 1.6cm]{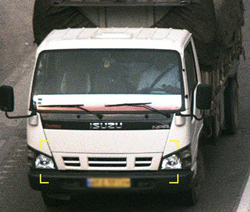}&
   \includegraphics[width = 1.6cm]{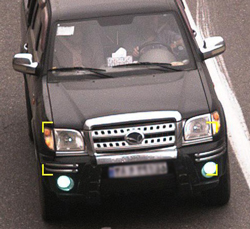}&
     \includegraphics[width = 1.6cm]{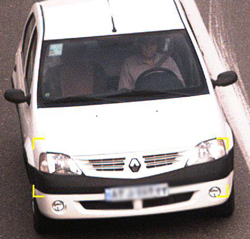}&
   \fcolorbox{brown}{white}{\includegraphics[width = 1.6cm]{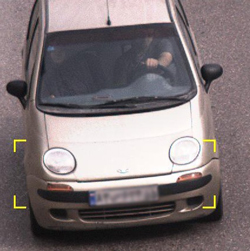}}&
\includegraphics[width = 1.6cm]{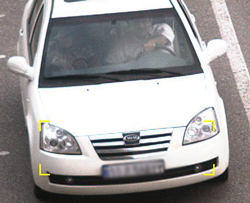}&
\fcolorbox{brown}{white}{\includegraphics[width = 1.6cm]{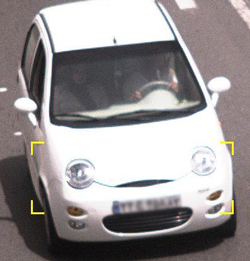}}&
\includegraphics[width = 1.6cm]{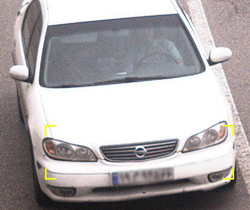}&
\includegraphics[width = 1.6cm]{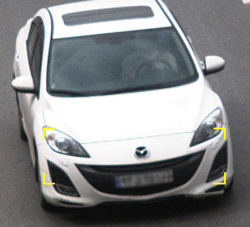}  \\
 C17&C18&C19&C20&C21&C22&C23&C24\vspace{-0cm}\\
\includegraphics[width = 1.6cm]{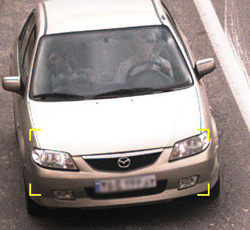}&
 \includegraphics[width = 1.6cm]{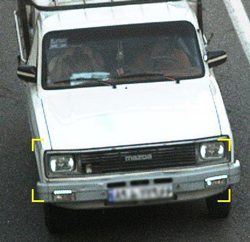} &
   \includegraphics[width = 1.6cm]{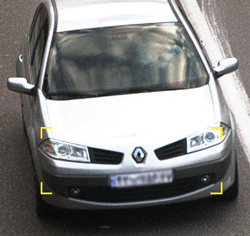}&
   \includegraphics[width = 1.6cm]{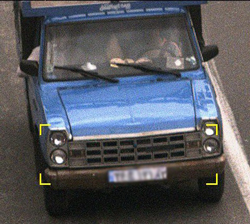} &
      \fcolorbox{green}{white}{\includegraphics[width = 1.6cm]{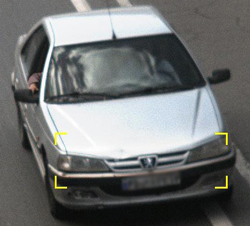}}&
   \includegraphics[width = 1.6cm]{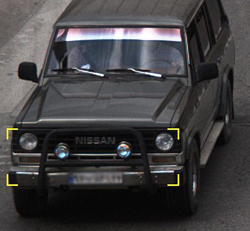}&

\fcolorbox{blue}{white}{\includegraphics[width = 1.6cm]{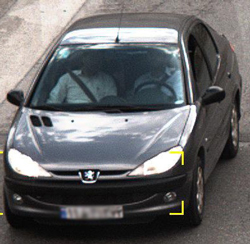}}&
\fcolorbox{green}{white}{\includegraphics[width = 1.6cm]{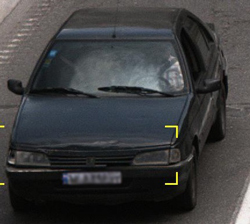}}\\
 C25&C26&C27&C28&C29&C30&C31&C32\vspace{-0cm}\\
\includegraphics[width = 1.6cm]{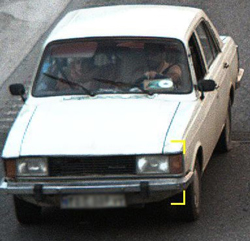}&
\includegraphics[width = 1.6cm]{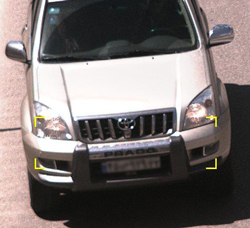}  &
\includegraphics[width = 1.6cm]{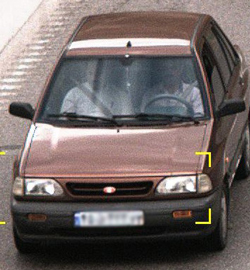}&
 \includegraphics[width = 1.6cm]{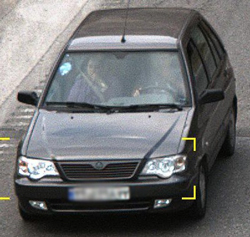}  &
 \includegraphics[width = 1.6cm]{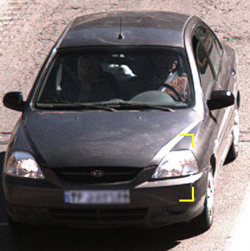}&
   \includegraphics[width = 1.6cm]{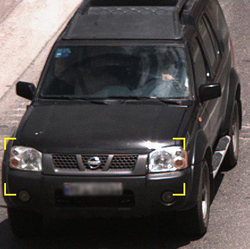} &
     \includegraphics[width = 1.6cm]{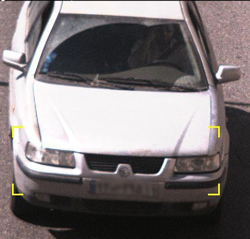}&
   \includegraphics[width = 1.6cm]{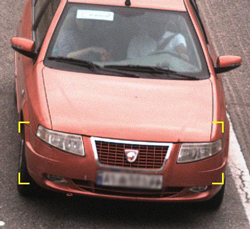}\\
 C33&C34&C35&C36&C37&C38&C39&C40\vspace{-0cm}\\
\includegraphics[width = 1.6cm]{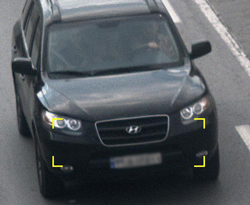}&
\includegraphics[width = 1.6cm]{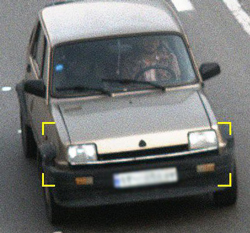}&
\includegraphics[width = 1.6cm]{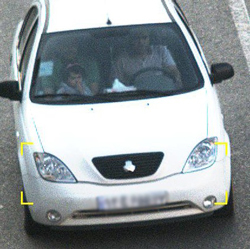}&
\includegraphics[width = 1.6cm]{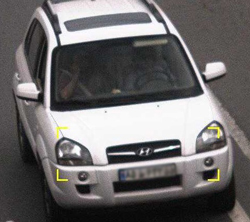}  &
\includegraphics[width = 1.6cm]{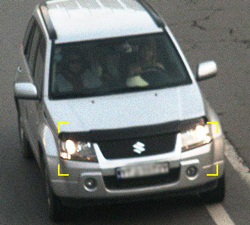}&
 \includegraphics[width = 1.6cm]{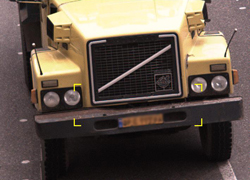}  &
   \fcolorbox{green}{white}{\includegraphics[width = 1.6cm]{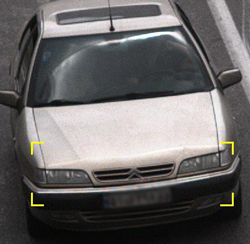}}&
   \includegraphics[width = 1.6cm]{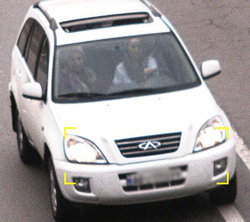}\\
    C41&C42&C43&C44&C45&C46&C47&C48\vspace{-0cm}\\
     \fcolorbox{blue}{white}{ \includegraphics[width = 1.6cm]{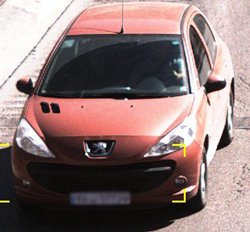}}&
   \includegraphics[width = 1.6cm]{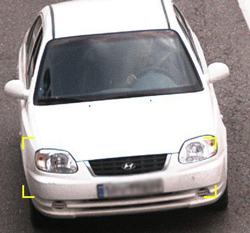}&
   \fcolorbox{white}{white}{\includegraphics[width = 1.6cm]{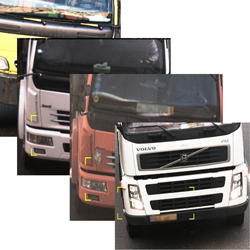}}&
   \fcolorbox{white}{white}{\includegraphics[width = 1.6cm]{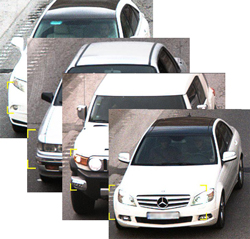}}
   \vspace{0.1cm}\\
 C49&C50&U1&U2\vspace{-0cm}\\
\end{tabular}
\caption{Example images of the Iranian on-road vehicles demonstrating the 50 classes. As seen the dataset contains a wide variety of vehicles which some of them are very similar. As an example, images of classes \{C29, C32, C47\} green border or \{C31, C49\} blue border or \{C20, C22\} brown border and \{C5, C6\} red border, are very similar which make the classification problem very challenging. Images \{U1, U2\} show two examples from the classes unknown heavy and unknown light respectively. The two unknown classes contain varied vehicles. Figure~\ref{fig:Unknowns} shows some example of two unknown classes. }
\label{fig:Objs}
\end{center}
\vspace{-0.7 cm}
\end{figure*}

\subsection{ Experimental Results}

\renewcommand{\arraystretch}{2}

\begin{table}[t]
\scriptsize
\centering
\caption{ The Accuracy of proposed method with and without confidence level computation step on the Iranian on-road vehicle dataset. Here C indicates the number of comparisons in LLC algorithm and W indicates the number of words in the dictionary. ORV-MMR-S1 stands for the framework when the prediction is done based on only the first classification step, while ORV-MMR-S2 represents the method when the classification accuracy based on the confidence level step.  }

\begin{tabular}{p{1 in}|p{0.4 in}|p{0.4 in}|p{0.4 in}|p{0.4 in}} 
	\hline \hline 
	Method & 1200W 100C & 1200W 500C & 3600W 100C & 3600W 500C\\ \hline 
	ORV-MMR-S1 & 96.60 \% & 97.05 \%  & 97.51 \%  & 98.52 \%  \\ 
	ORV-MMR-S2 & 96.06 \%  & 96.13 \% & 96.28 \%  & 97.56 \%  \\ 

 \hline \hline
\end{tabular}
\newline
\newline
 \label{AccuracyTable19}
\end{table}

\begin{table}[t]
	\scriptsize
	\centering
	\caption{ The running time of proposed method with and without confidence level computation step, with different dictionary size and different number of comparison for kd-tree structure of LLC method on the Iranian on-road vehicle dataset. The  reported times is in second. }
	\begin{tabular}{p{1.4in}|p{0.3 in}|p{0.3 in}|p{0.3 in}|p{0.3 in}} 
		\hline \hline 
		Method & 1200W 100C & 1200W 500C & 3600W 100C & 3600W 500C\\  \hline \hline
		ORV-MMR-S1 & 0.0709  & 0.0989  & 0.0912  & 0.1115  \\ 
		ORV-MMR-S2 & 0.078  & 0.1090  & 0.0920  & 0.1221  \\ 
		
		\hline \hline 
	\end{tabular}
	\newline
	\newline
	\label{runningtime}
\end{table}


As the first experiment, the effect of different set of parameters in the proposed  ORV-MMR framework is examined. As mentioned in Section~\ref{sec:MSC}, a multi-stage classification technique is designed to compute the confidence level while the  class label of the vehicle is being predicted. However, it is possible to predict the class label without applying  the second stage and the main role of the second stage is more about  computing the confidence of the classification. Therefore, the proposed framework is evaluated with two different approaches: I) \mbox{ORV-MMR-S1} which  classifies the images  with only the first stage framework and II) \mbox{ORV-MMR-S2} which includes first and second stages for predicting the class label of the input images.

Table~\ref{AccuracyTable19} shows  the effects of different dictionary sizes on the accuracy of proposed methods (\mbox{ORV-MMR-S1} and \mbox{ORV-MMR-S2}) on Iranian on-road dataset. As  seen, increasing the size of dictionary improves the accuracy.The larger dictionary size generate a larger feature set which provides a more accurate model.  Table~\ref{AccuracyTable19} also shows that increasing the maximum comparison of kd-tree structure improves the modeling accuracy. The reported results demonstrate that adding an extra NN layer (the second stage of ORV-MMR-S2 ) does not considerably change the accuracy of the model. However, the results of Table~\ref{AccuracyTable19} convince us to use the first stage's output for classification. The advantage of using the second stage is to provide  the label's confidence.

Table~\ref{runningtime} demonstrates the computational complexity of both proposed frameworks in terms of the running time. The reported results are achieved using a personal computer with a 3.4 GHz Intel CPU and 32 Gigabytes RAM. The average running time is reported based on 100 randomly selected images from Iranian on-road vehicle dataset. As shown in Table~\ref{runningtime}, the maximum comparison of kd-tree structure and different sizes of dictionary directly affect the running time of the proposed methods. Here we compare 100 and 500 maximum comparisons in the kd-tree structure and two different dictionary sizes \{1200,3600\}. Table~\ref{runningtime} also shows that the effect of  NN layer (i.e., the second stage in ORV-MMR-S2) is negligible on  running time of the framework and providing the confidence level does not considerably increase the computational complexity of the framework. 
The significant impact of adding the NN layer to the proposed framework is  the improvement in  robustness of the framework  and providing a meaningful confidence for unknown classes. 

Tables \ref{AccuracyTable19} and \ref{runningtime} demonstrate that the most appropriate method for a real-time MMR system is a ORV-MMR-S2 which benefits from the first stage for classification and second stage for providing a meaningful confidence. In other words, a multi-stage classification model consist of fast-LLC, SVM and NN, with a dictionary size of 3600 words and a kd-tree with maximum 100 comparisons to find the nearest neighbor is selected. Therefore, ORV-MMR-S2 with the aforementioned configuration is utilized to compare with the state-of-the-art methods.

\begin{figure}[!t]
\centering
\includegraphics[trim={1cm 3.5cm 1cm 0},clip,width=1\linewidth]{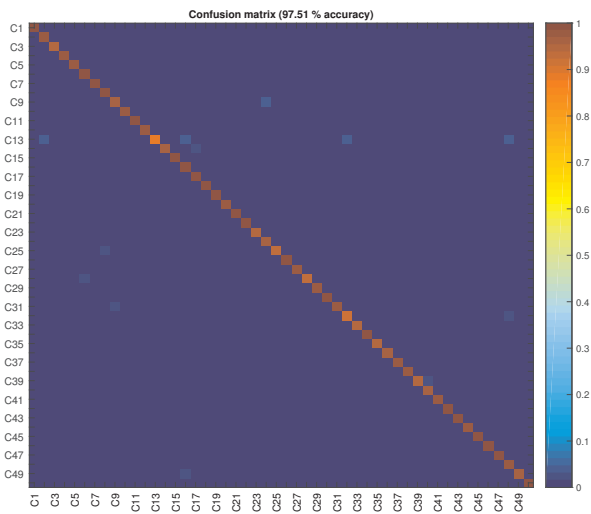}
\caption{Confusion matrix of the proposed method on the 50 known classes. As seen, misclassification occurs on the classes $\{C39,C40\}$ which are the same vehicle model. Additionally this figure shows that the most misclassification rate  belongs to $C13$. Odd class label numbers are only shown for the better visualization purposes.}
\label{fig:image9}
\end{figure}

Figure~\ref{fig:image9} demonstrates the confusion matrix of 50 known classes  of Iranian on-road vehicle dataset predicted by the proposed method. As seen, the overall accuracy of the proposed method is 97.51\%. Figure~\ref{fig:image9}  shows that the misclassification  occurs on the classes $\{C39,C40\}$ such that they can be classified in the same group as they belong to a unique make and model vehicle but with tiny difference. This is evident by comparing them in Figure\ref{fig:Objs}. Additionally, Figure~\ref{fig:image9} shows that class $C13$ has the lowest accuracy among other classes since it has the least number of training samples.

\begin{figure}[!t]
	\centering
	\includegraphics[trim={1cm 3.5cm 1cm 0},clip,width=1\linewidth]{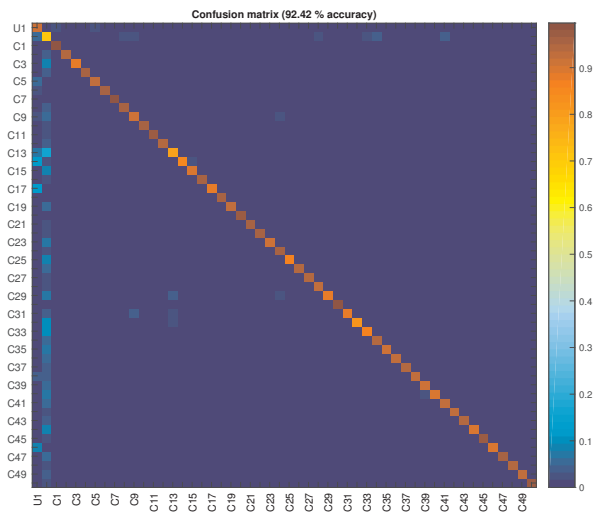} 
	\caption{Confusion matrix of the proposed methods on 52 classes of Iranian on-road vehicle dataset. The firs two classes are U1 and U2.  As  seen the classes U1 and U2 cause many false positive samples. Odd class label numbers are only shown for  better visualization purposes.}
	\label{fig:image8}
\end{figure}

Figure~\ref{fig:image8} demonstrates the confusion matrix of combined 50 vehicle classes and two extra unknown classes on the Iranian on-road vehicle dataset. The reported accuracy for  52 classes is 92.42\%. This result shows, due to the large within class variances of unknown classes, specially the light unknown vehicles, the average accuracy of these two classes (81\%) is lower than average accuracy of all other classes. In addition, as it was demonstrated in Figure~\ref{fig:Data_52}, some training classes such as $\{C13, C14, C15\}$ have much less number of samples than others such that the models of these classes could not be trained very well. In other words, unbalanced training data also makes the problem hard which has a negative impact on the accuracy. However, the one versus all SVM method is obtained to relax the effect of large between class variance of the vehicles and unbalanced dataset. As mentioned before, there are several similar classes such as  bus, track, van and sedan which must be classified as separate labels imposing a large within class variance in to the model.

\begin{figure}[t]
	\centering
	\vspace{0.1cm}
	
	\begin{tabular}{l||l|cccc}
		\centering
		\multirow{2}{*}{\raisebox{-15 pt}[0 pt][0 pt]{\rotatebox{90}{\footnotesize Input Image}}}& 
		 \raisebox{0 pt}[0 pt][0 pt]{\rotatebox{90}{\tiny Input Image}}& 
		\includegraphics[width = 1.2cm]{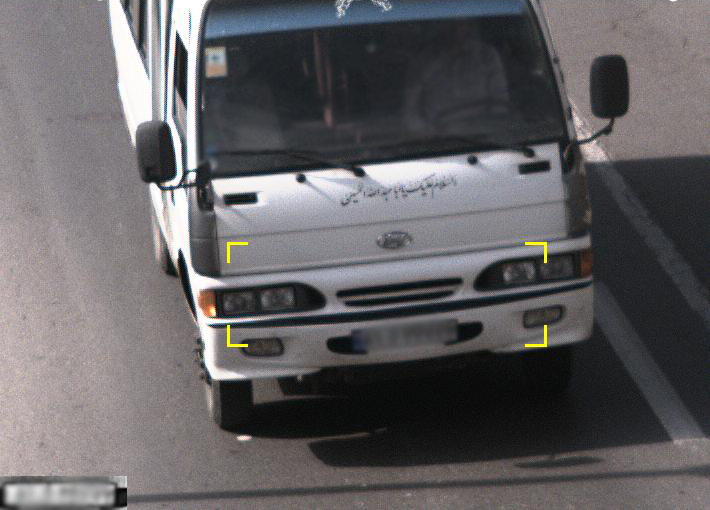}&
		\includegraphics[width = 1.2cm]{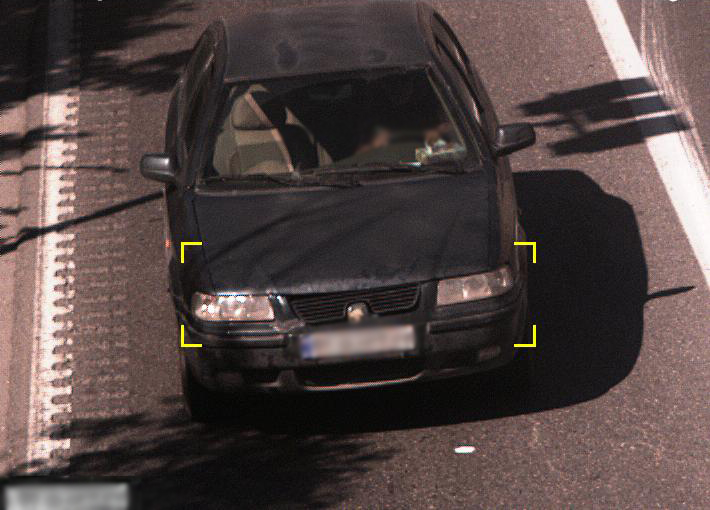}&
		\includegraphics[width = 1.2cm]{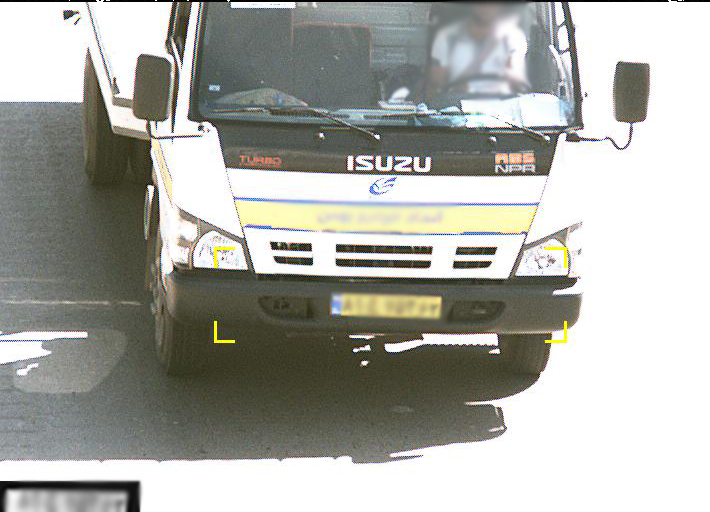}  &
		\includegraphics[width = 1.2cm]{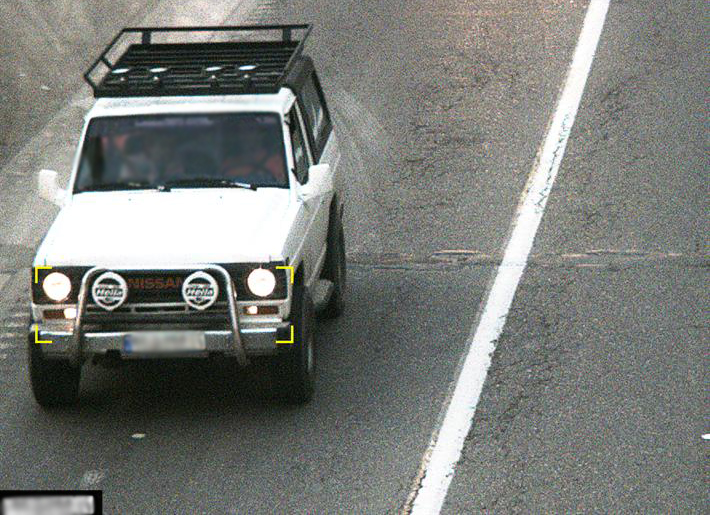} \\
	   &\raisebox{0 pt}[0 pt][0 pt]{\rotatebox{90}{\tiny GT}} &C15&C39&C17&C30 \\\hline \hline
	   
	   	\multirow{2}{*}{\raisebox{-10 pt}[0 pt][0 pt]{\rotatebox{90}{ \footnotesize ORV-MMR}}}& 	
		\raisebox{0 pt}[0 pt][0 pt]{\rotatebox{90}{\tiny Sample-Img}} &
		\includegraphics[width = 1.2cm]{Pic/c15}&
		\includegraphics[width = 1.2cm]{Pic/c39}&
		\includegraphics[width = 1.2cm]{Pic/c17}&
		\includegraphics[width = 1.2cm]{Pic/c30}\\
		&\raisebox{-3 pt}[0 pt][0 pt]{\rotatebox{90}{\tiny Pred. Cls.}} &C15&C39&C17&C30\\\hline	
		
		\multirow{2}{*}{\raisebox{-20 pt}[0 pt][0 pt]{\rotatebox{90}{ \footnotesize F-G-VMMR~\cite{IEEEhowto:Fang}}}}& 	
		\raisebox{3 pt}[0 pt][0 pt]{\rotatebox{90}{\tiny Sample-Img}} &
		\includegraphics[width = 1.2cm]{Pic/c14}&
		\includegraphics[width = 1.2cm]{Pic/c10}&
		\includegraphics[width = 1.2cm]{Pic/c11}&
		\includegraphics[width = 1.2cm]{Pic/c38} \\
		&\raisebox{-3 pt}[0 pt][0 pt]{\rotatebox{90}{\tiny Pred. Cls.}}&C14&C10&C11&C38\\\hline			
		
		\multirow{2}{*}{\raisebox{-10 pt}[0 pt][0 pt]{\rotatebox{90}{\footnotesize AlexNet~\cite{IEEEhowto:AlexNet}}}}& 	
		\raisebox{0 pt}[0 pt][0 pt]{\rotatebox{90}{\tiny Sample-Img}} &
		\includegraphics[width = 1.2cm]{Pic/c14}&
		\includegraphics[width = 1.2cm]{Pic/c45}&
		\includegraphics[width = 1.2cm]{Pic/c11}&
		\includegraphics[width = 1.2cm]{Pic/c38}\\
         &\raisebox{-9  pt}[0 pt][0 pt]{\rotatebox{90}{ \tiny Pred. Cls.}}	 &C14&C45&C11&C3
	\end{tabular}

	\caption{Some examples of Iranian on-road dataset which were recognized correctly by the proposed method while the competing  methods could not classified them correctly. The first row shows the input images and their labels. Other rows show the results of the proposed method, F-G-VMMR and AlexNet.  The image examples of predicted class labels (Pred. Cls. rows) are shown in Sample-Img rows for comparison purposes.  }
	\label{fig:Wrong}
\end{figure}

Table~\ref{AccuracyPerClass} shows the per-class accuracy of the proposed method, F-G-VMMR proposed by Fang  {\it et al}~\cite{IEEEhowto:Fang} and AlexNet on Iranian on-road vehicle dataset. As seen, the proposed method outperforms two other methods on 38 out of 50 tested class labels. It is worth to note the proposed method ORV-MMR outperforms two other competing methods on the average class accuracy as well. The proposed method provides a much higher accuracy in classes $\{C15,C17,C30,C35\}$ compared to other methods. Figure~\ref{fig:Wrong} shows the example images where  the proposed method could classify correctly while they are misclassified by two other competing algorithms. These results  illustrate that when the input images are corrupted by shadows or comes with different illuminations (as evident by class $C35$ in the Figure~\ref{fig:Wrong})  a more robust feature descriptor is required to recognize the input image. Another two samples from classes $\{C15,C17\}$ show that CNN feature fails the classification when some changes are added to vehicles or the vehicle images are rotated while the proposed ORV-MMR approach could classify them correctly as it obtains robust feature descriptors.

Table~\ref{AccuracyPerClass} demonstrates that the proposed method outperforms AlexNet in classification which can be justified by the the small size of training dataset and challenging input images. It also should be considered that  the AlexNet architecture could not handle  the small within class variances of the classes in the available dataset. However to mitigate these issues  data augmentation is used to increase the number of training data. Here we used flipping technique to do the augmentation.

\begin{figure}
	\vspace{- 0.1 cm}
	\fboxsep=1mm
	\fboxrule=2pt
	\begin{center}
		\begin{tabular}{cccc}
			\fcolorbox{black}{white}{\includegraphics[width = 1.4cm]{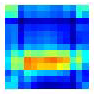}}&
			\fcolorbox{black}{white}{\includegraphics[width = 1.4cm]{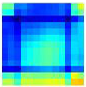}}&
			\fcolorbox{orange}{white}{\includegraphics[width = 1.4cm]{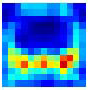}}&
			\fcolorbox{orange}{white}{\includegraphics[width = 1.4cm]{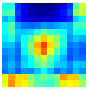}}
			\vspace{- 0.2cm}\\
			IO-Whole&IO-Head Part&CC-Whole&CC-Head Part\vspace{-0cm}\\
		\end{tabular}
		\caption{ The heatmaps extracted from all training data of Iranian On-road (IO) and CompuCar(CC) Dataset in two hierarchical levels by using the method of ~\cite{IEEEhowto:Fang}. The first two heatmaps (black border) show the heatmap which are extracted from all training data of whole image (IO-Whole) and selected head part (IO-Head Part) of Iranian on-road dataset respectively. The last two heatmaps (orange border) show the heatmap which are extracted from the whole image (CC-Whole) and the selected headpart (CC-Head Part) of CompuCar dataset. Comparing the heatmaps extracted from Iranian on-road dataset and Compucar illustrates that the vehicle images must be aligned for the F-G-VMMR framework to be able to find the discriminative features. The method could find discriminative features (head part and light areas) in the CompuCar dataset (right heatmaps) while it could not find discriminative features in Iraninain On-road dataset (left heatmaps).}
		\vspace{0.5 cm}
		\label{fig:HeatMapObjs-small}
	\end{center}
	\vspace{-0.7 cm}
\end{figure}

F-G-VMMR framework proposed by Fang {\it et al.}~\cite{IEEEhowto:Fang} also could not provide higher modeling accuracy compared to the proposed framework. The hierarchical part detection designed in F-G-VMMR fails on  the Iranian on-road vehicle dataset since this dataset contains various type of car such as bus, mini-bus, track, mini-track in addition to sedan in different scale and field of views. Additionally, the input images of the dataset are not aligned which makes the classification more difficult for F-G-VMMR part detection algorithm. Hierarchical part detection  is implemented by use of a CNN which is trained on  all available training data. They extract the heatmap which is the pixel value of the last convolutional layer of each training data and average map is generated by averaging all heatmaps of training images. The higher value of average heatmap pixels producing  brighter color pixels and consequently the regions containing such pixels are more discriminative for recognition compared to regions containing darker pixels in average heatmap. This procedure is performed on the selected region of input images of training data and continues till no new region is detected.

 In our dataset, their procedure stopped in the first step (a head part is detected). Additionally, the selected head part does not work for all vehicle classes such as buses, mini-buses, tracks and mini-tracks since their head parts are different in size and form in compare to sedans. Figure \ref{fig:HeatMapObjs-small} compares the results of their hierarchical part detection on both used dataset.  As seen, F-G-VMMR framework mainly selected the head-part to extract discriminative features.


\begin{table}[t]
	\scriptsize
	\centering
	\caption{ The accuracy of all correctly recognized samples of the best and the most appropriate configurations of the proposed method on the 50 classes of Iraninan on-road vehicle dataset and the CompuCar dataset. The proposed framework is compared with AlexNet architecture and F-G-VMMR~\cite{IEEEhowto:Fang} frameworks. }
	\begin{tabular}{p{1.8in}|p{0.6 in}|p{0.6 in}} 
		\hline \hline 
		Method & Iranian on-road  & CompuCar \\ \hline \hline
	ORV-MMR-S2& \textbf{97.51}
		\% & 98.41 \% \\ 
		F-G-VMMR~\cite{IEEEhowto:Fang}  & 95.26 \% & \textbf{98.63} \% \\ 
		AlexNet & 94.14 \% & 94.31 \% \\
		\hline \hline
	\end{tabular}
	\newline
	\label{AllAccuracy}
	
\end{table}

Result in Table~\ref{AllAccuracy} shows that  F-G-VMMR method fails on Iranian on-road dataset since their part detection methods can not detect the discriminative regions  in some categories of Iranian on-road dataset. As a result, the CNN features are not robust enough when the input images are taken in a different illumination or with different field  of views captured by different cameras. Table~\ref{RunningTimeTwoPaper} reports the running-time of the competing methods on Iranian on-road dataset images. The proposed ORV-MMR framework provide faster computation compared to F-G-VMMR approach which makes it more desirable for real-time applications.  
\renewcommand{\arraystretch}{2}
\begin{table*}[t]
	\scriptsize
	\centering
	\caption{ The per-class Accuracy of the proposed method and two other methods on Iranian on-road vehicle dataset. The last column shows the mean average precision of vehicle classes (i.e., per-class mean accuracy).}
	\centering
	\resizebox{\textwidth}{!}
	{\begin{tabular}{lcccccccccccccccccc}
		 \hline \hline 
		 Method & C1&C2&C3&C4&C5&C6&C7&C8&C9&C10&C11&C12&C13&C14&C15&C16&C17&C18 \\ \hline 
		 ORV-MMR-S2 & \textbf{99.77\%} &  \textbf{98.04\%}&   \textbf{95.25\%}&   \textbf{97.55\%}&   \textbf{98.15\%}&   \textbf{98.51\%}&  \textbf{100\%}&   98.65\%&   \textbf{96.67\%}&   \textbf{97.90\%}&   \textbf{98.82\%}& \textbf{97.73\%}&   87.50\%&   95.31\%&   \textbf{98.52\%}&   \textbf{99.72\%}&   \textbf{99.67\%}&  \textbf{100\%}\\
		 F-G-VMMR~\cite{IEEEhowto:Fang}& 99.32\%&  97.55\%&   94.30\%&   98.04\%&   96.91\%&   97.01\%&  \textbf{100\%}&   97.97\%&   91.67\%&   93.01\%&   97.65\%& 95.45\%&   90.62\%&   93.75\%&   91.11\%&   95.56\%&   95.69\%&   97.79\% \\
		 AlexNet & \textbf{99.77\%}&   96.08\%&   91.77\%&   \textbf{97.55\%}&   96.91\%&   \textbf{98.51\%}&   99.59\%&  \textbf{100\%}&   93.33\%&   90.21\%&   \textbf{98.82\%}& 93.18\%&   \textbf{93.75\%}&   \textbf{96.87\%}&   85.19\%&   95.56\%&   93.38\%&   99.26\% \\
		\hline \hline
		 Method &C19&C20& C21&C22&C23&C24&C25&C26&C27&C28&C29&C30&C31&C32&C33&C34&C35&C36 \\ \hline 
		 ORV-MMR-S2 & \textbf{99.25\%} &   \textbf{97.83\%} & \textbf{99.48\%} & \textbf{98.57\%} & \textbf{94.89\%} & 96.61\%&   \textbf{93.02\%} & \textbf{100\%}&   97.64\%&  \textbf{93.56\%} &   \textbf{97.26\%}&  \textbf{100\%} & \textbf{97.15\%}&  \textbf{91.67\%}&   \textbf{95.1\%}&  \textbf{100\%}& \textbf{94.28\%}&   \textbf{96.29\%}  \\ 
		 F-G-VMMR~\cite{IEEEhowto:Fang} &  97.01\%&   97.10\%&  98.43\%&   92.86\%& \textbf{94.89\%} &   95.76\%&   92.25\%&   94.44\%&   97.64\%&   90.99\%&   90.87\%&   96.57\%&   92.88\%&   89.44\%&   89.22\%&  97.93\%&   90.25\%&   95.85\% \\
		 AlexNet & 94.78\%&   95.65\%&   95.81\%&   95.71\%& 93.43\%&   \textbf{98.31\%}&   89.92\%&   94.44\%&  \textbf{100\%}&   86.69\%&   88.58\%&   93.84\%&   87.18\%&   82.78\%&   90.2\%&  97.93 \%&   90.47\%&   91.27\%
		 
		 \\ \hline \hline
		 Method &C37&38&C39&C40&C41&C42&C43&C44&C45&C46&C47&C48&C49&C50&\multicolumn{3}{c}{Average Per-Class Accuracy}\\ \hline
		 ORV-MMR-S2 & \textbf{97.81\%} & \textbf{97.62\%}& \textbf{94.72\%} & \textbf{95.36\%}& 97.15\%&  91.67\%&   95.1\%&  \textbf{100\%}&   94.28\%&   96.29\%&   \textbf{97.81\%}&   \textbf{97.62\%}&   94.72\%&   95.36\% &&\textbf{97.41\%}
		 \\
		 F-G-VMMR~\cite{IEEEhowto:Fang} & 95.63\%&   95.24\%&   84.15\%&   \textbf{95.36\%} &   \textbf{98.49\%} & \textbf{98.50\%}&   \textbf{98.65\%} &   93.91\%& \textbf{98.12\%} &   98.97\%&   97.54\%&   96.25\%& \textbf{97.7\%} &  \textbf{96.77\%} &&95.26\%
		 \\
		 
		 AlexNet & 94.54\%&   95.24\%& 86.18\%&   93.38\%&   97.49\%&   \textbf{98.50\%}&   96.62\%&   94.35\%& \textbf{98.12\%}&  \textbf{99.32\%}&   96.72\%&   96.25\%&   94.47\%&   94.01\%&&94.43\%
		 
		 \\
		  \hline \hline	
		 	
		\end{tabular}}
 \label{AccuracyPerClass}
	\end{table*}

\begin{table}[t]
\scriptsize
\centering
\caption{The running time of the proposed methods on Iranian on-road dataset. The running time is reported based on only the classification step. Maximum 100 comparison are used in LLC algorithm. The unit scale is second. }
\begin{tabular}{p{1.2 in}|p{0.8 in}|p{0.9 in}} 
 \hline \hline 
Method & Programming language & Iranian on-road dataset \\ \hline \hline
ORV-MMR-S2 & Matlab & 0.09 \\
F-G-VMMR~\cite{IEEEhowto:Fang}  & Matlab & 0.12 \\
AlexNet & Matlab & 0.06 \\
 \hline \hline 
\end{tabular}
 \label{RunningTimeTwoPaper}
 \newline
\end{table}

\section{ Conclusion}
We proposed a new real-time framework  for vehicle make and model recognition which address the problem via a fine-grained classification approach. The proposed on-road vehicle make and model (ORV-MMR) framework provides a great modeling accuracy on 50 different class of Iranian on-road vehicle dataset.  To provide a comprehensive framework for recognizing any vehicle, we proposed two extra classes (unknown heavy and unknown light) where any vehicle does not belong to 50  classes, is classified as one of these two unknown class labels.
  
 The proposed framework provides a strong theoretical background along with a good generalization and robustness based on classification accuracy. Additionally, the small number of parameters and its easy fine-tuning advantage make the proposed framework strong dealing with small number of training data of an unbalance dataset . The proposed framework can be divided into two independent stages where the first stage predicts the class label while the second stage provides a confidence level for the classification.  The proposed ORV-MMR framework was compared with two state-of-the-art methods which outperforms both methods on Iranian on-road dataset while obtains comparable accuracy in CompuCar dataset.    

\section*{Acknowledgment}
We would like to thank the NAJA Traffic Research Center, for providing the Iranian on-road vehicles dataset and their supports.


\ifCLASSOPTIONcaptionsoff
  \newpage
\fi

\end{document}